\documentclass{article}

\usepackage[preprint]{neurips_2023}

\usepackage[utf8]{inputenc} % allow utf-8 input
\usepackage[T1]{fontenc}    % use 8-bit T1 fonts

\usepackage{graphicx}
\usepackage{amsmath}
\usepackage{amssymb}
\usepackage{tabularx}
\usepackage{cite}
\usepackage{amsfonts}
\usepackage{algorithmic}
\usepackage{subcaption}
\usepackage{multirow}
\usepackage{booktabs}       % professional-quality tables

\usepackage{bm}
\usepackage{color}
\usepackage{xcolor}
\usepackage{algorithm}
\usepackage{xspace}
\usepackage{soul}

\usepackage{enumitem}
\usepackage{textcomp}  % Required for encoding \textbigcircle
\usepackage{scalerel}  % Required for emoji \scalerel

\usepackage[colorlinks,linkcolor=blue]{hyperref}

\def\logo{\scalerel*{\includegraphics{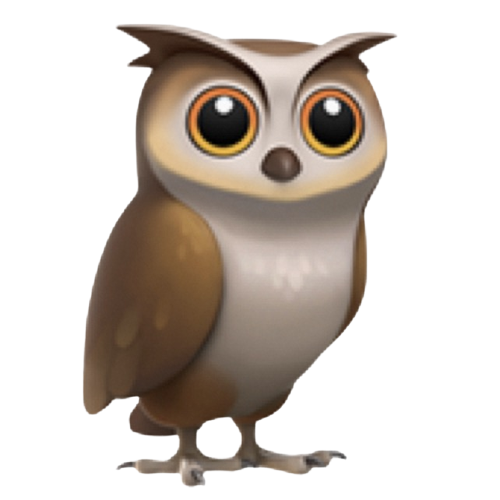}}{\textrm{\textbigcircle}}}
\newcommand{\titledmodelname}{mPLUG-Owl\logo\xspace}
\newcommand{\modelname}{mPLUG-Owl\xspace}
\newcommand{\evalsetname}{OwlEval\xspace}

\title{\titledmodelname: Modularization Empowers Large Language Models with Multimodality}

\author{%
  Qinghao Ye\thanks{Equal contribution}\hspace{1.5mm}, Haiyang Xu$^*$, Guohai Xu, Jiabo Ye, Ming Yan\thanks{Corresponding author}, Yiyang Zhou, \\ \textbf{Junyang Wang, Anwen Hu, Pengcheng Shi, Yaya Shi, Chenliang Li, Yuanhong Xu,} \\ \textbf{Hehong Chen, Junfeng Tian, Qi Qian, Ji Zhang, Fei Huang, Jingren Zhou} \\
DAMO Academy, Alibaba Group \\
{\small \texttt{\{yeqinghao.yqh, shuofeng.xhy, guohai.xgh, ym119608\}@alibaba-inc.com}} 
}

\begin{document}

\maketitle

\begin{abstract}
Large language models (LLMs) have demonstrated impressive zero-shot abilities on a variety of open-ended tasks, while recent research has also explored the use of LLMs for multi-modal generation. In this study, we introduce \textbf{\modelname}, a novel training paradigm that equips LLMs with multi-modal abilities through modularized learning of foundation LLM, a visual knowledge module, and a visual abstractor module. This approach can support multiple modalities and facilitate diverse unimodal and multimodal abilities through modality collaboration. The training paradigm of \modelname involves a two-stage method for aligning image and text, which learns visual knowledge with the assistance of LLM, while maintaining and even improving the generation abilities of LLM. In the first stage, the visual knowledge module and abstractor module are trained with frozen LLM module to align the image and text. In the second stage, language-only and multi-modal supervised datasets are used to jointly fine-tune a low-rank adaption (LoRA) module on LLM and the abstractor module by freezing the visual knowledge module. We carefully build a visually-related instruction evaluation set \textbf{\evalsetname}. Experimental results show that our model outperform existing multi-modal models,  demonstrating \modelname's impressive instruction and visual understanding ability, multi-turn conversation ability and knowledge reasoning ability. Besides, we observe some unexpected and exciting abilities such as multi-image correlation and scene text understanding, which makes it possible to leverage it for harder real scenarios, such as vision-only document comprehension.
Our code, pre-trained model,  instruction-tuned models, and evaluation set are available at \href{https://github.com/X-PLUG/mPLUG-Owl}{https://github.com/X-PLUG/mPLUG-Owl}. The online demo is available at \href{https://www.modelscope.cn/studios/damo/mPLUG-Owl/}{https://www.modelscope.cn/studios/damo/mPLUG-Owl}.
\end{abstract}

% This approach stimulates a broad range of unimodal and multimodal abilities.

\section{Introduction}
Large language models (LLMs) such as GPT-3 \citep{gpt3}, BLOOM \citep{bloom}, LLaMA \citep{llama} have experienced rapid development to make general artificial intelligence possible, which demonstrates impressive zero-shot abilities on various linguistic applications. However, except GPT-4 \citep{gpt4}, current general LLMs cannot support different modalities of input and develop impressive multimodal abilities. 

Although GPT-4 \citep{gpt4} has exhibited remarkable multimodal abilities, the methods behind its extraordinary abilities remain a mystery. Recently, researchers have been extending LLMs to understand visual inputs in two different paradigms: systematic collaboration and end-to-end trained models. However, systematic collaboration approaches, including Visual ChatGPT \citep{visualchatgpt}, MM-REACT \citep{mmreact}, and HuggingGPT \citep{hugginggpt}, are designed to facilitate the coordination of various vision models or tools to express visual information with text descriptions. However, these approaches may not be able to comprehend specific multimodal instructions due to their lack of alignment with different modalities. Additionally, these approaches may encounter challenges related to inference efficiency and cost. End-to-end models, such as BLIP-2 \citep{blip2}, LLaVA \citep{llava}, and MiniGPT-4 \citep{minigpt4} aim to use unified models to support different modalities. However, these models have some limitations as they take frozen visual models, which may lead to inadequate alignment due to the limited number of parameters. Moreover, they cannot unlock various abilities due to missing unimodal and multimodal instruction.

In this paper, we present \modelname with an innovative modularized training paradigm for large multi-modal language models that can support multiple modalities concurrently, drawing inspiration from the concept of modularization \citep{mplug2, mplug, e2evlp, hitea}. Our method harnesses the power of pre-trained LLM, visual knowledge module, and connected visual abstractor module to achieve effective alignment between images and text, and utilizes a two-stage training scheme to stimulate impressive unimodal and multimodal abilities. Our approach even enhances the strong generation abilities of LLM by modality collaboration between modalities. In the first step, we align the image and text to acquire comprehensive visual knowledge using text-image pairs, which is accomplished by training the  visual knowledge module and abstractor module with the frozen LLM module. Subsequently, we fine-tune mPLUG-Owl with language-only and multi-modal instructions to unlock a range of unimodal and multimodal abilities. We freeze the visual knowledge module and train low-rank adaption (LoRA) \citep{lora} on LLM and visual abstractor module jointly. This approach allows for the effective integration of textual and visual information, facilitating the development of versatile and robust cognitive abilities.

Our experiments on a carefully-built visually related instruction evaluation set \evalsetname shows that \modelname outperforms existing models such as MiniGPT-4 \citep{minigpt4} and LLaVA \citep{llava}. We separately verifies 
\modelname's remarkable abilities in instruction understanding, visual understanding, knowledge transfer, and multi-turn dialogue. %Knowledge transfer ability combines instructional and visual information to provide additional knowledge support. Multi-turn dialogue enables clear referencing and understanding of contextual semantics. 
Abundant ablation study is performed to show the effectiveness of our training paradigm. Furthermore, we find some unexpected emerging ability such as multi-image correlation, multilingual conversation and scene text understanding.

%Moreover, we also verify that introducing multi-modal data during instruction tuning could further improve LLM’s performance on text-only tasks.

Our main contributions can be highlighted as follows:
\begin{itemize}
    \item We propose \modelname, a novel  training paradigm for large language models through modularization.
    \item We carefully construct an instruction evaluation set, dubbed \textbf{OwlEval}, to assess the capabilities of different models in the context of visual-related tasks.
    \item Experimental results demonstrate that \modelname excels in multi-modal instruction understanding and multi-turn dialogue, surpassing the performance of existing models.
\end{itemize}

\section{Related Work}
\subsection{Large Language Models}
In recent times, Large Language Models (LLMs) have garnered increasing attention for their exceptional performance in diverse natural language processing (NLP) tasks. Initially, transformer models such as BERT \citep{bert}, GPT \citep{gpt1}, and T5 \citep{t5} were developed with different pre-training objectives. However, the emergence of GPT-3 \citep{gpt3}, which scales up the number of model parameters and data size, showcases significant zero-shot generalization abilities, enabling them to perform commendably on previously unseen tasks. Consequently, numerous LLMs such as OPT \citep{opt}, BLOOM \citep{bloom}, PaLM \citep{palm}, and LLaMA \citep{llama} are created, ushering in the success of LLMs. Additionally, Ouyang et al. \citep{instructgpt} propose InstructGPT by aligning human instruction and feedback with GPT-3. Furthermore, it has been applied to ChatGPT \citep{chatgpt}, which facilitates conversational interaction with humans by responding to a broad range of diverse and intricate queries and instructions.

\subsection{Multi-Modal Large Language Models}
Despite the successful applications of LLMs in natural language processing, it is still struggling for LLMs to perceive other modalities such as vision and audio. Recently, researchers have been extending language models to understand visual inputs in two different paradigms: systematic collaboration and end-to-end trained models. Systematic collaboration approaches, such as Visual ChatGPT \citep{visualchatgpt}, MM-REACT \citep{mmreact}, and HuggingGPT \citep{hugginggpt}, leverage various vision experts or tools to express visual information with text descriptions. Subsequently, large language models, such as ChatGPT, can act as the agents, and be prompted to select the appropriate experts and tools for visual understanding. Finally, LLMs would summarize the output of these experts to answer user queries. 
% On the other hand, some approaches \citep{} treat language models as decoders in the vision-language model, which enables the model to share knowledge between the language and vision domains.
On the other hand, some approaches \citep{blip2, flamingo, llava} leverage the pre-trained large language model to build unified models for multi-modality.
For example, Flamingo \citep{flamingo} freezes the pre-trained vision encoder and large language model and fuses vision and language modalities with gated cross-attention showing impressive few-shot capabilities. Additionally, BLIP-2 \citep{blip2} designs Q-Former to align the visual features from the frozen visual encoder and large language models with Flan-T5 \citep{flant5} and OPT \citep{opt}. Moreover, PaLM-E \citep{palm-e} directly inputs features from sensor modalities with PaLM \citep{palm}, which has 520 billion parameters, contributing to robust performance in real-world perceptions. Furthermore, some powerful instruction-tuned language models that built upon open-sourced foundation model LLaMA \citep{llama}, such as Alpaca \citep{alpaca} and Vicuna \citep{vicuna}, exhibit comparable performance to ChatGPT \citep{chatgpt} and GPT-4 \citep{gpt4}. MiniGPT-4 \citep{minigpt4} and LLaVA \citep{llava} align these finetuned models with extracted visual features from the frozen visual backbone. In contrast, \modelname not only aligns the representation between the vision and language foundation model (e.g. CLIP and LLaMA) in terms of knowledge acquisition and grounding to the real world but also can understand language and multi-modal instructions, showcasing strong zero-shot generalization and multi-turn conversation capabilities.

\section{mPLUG-Owl}
\begin{figure}[!ht]
    \centering
    \includegraphics[width=\textwidth]{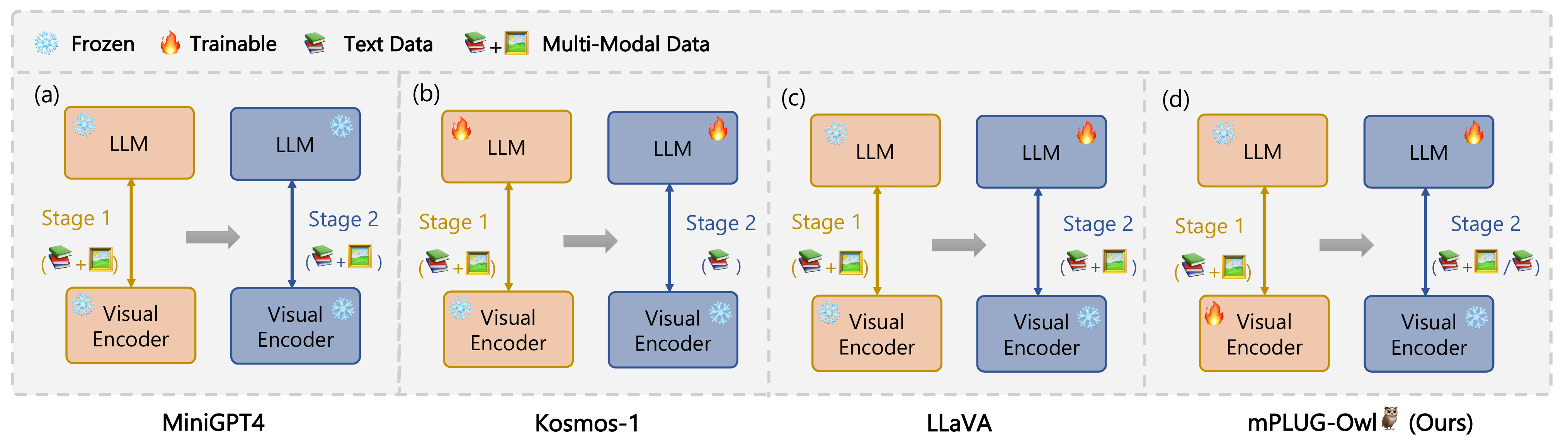}
    \caption{Comparison between different training paradigms. All of these methods are trained in a two-stage fashion. Stage 1 stands for pre-training and Stage 2 represents instruction tuning.}
    \label{fig:compare_method}
    \vspace{-2mm}
\end{figure}

\begin{figure*}[!ht]
    \centering
    \includegraphics[width=\textwidth]{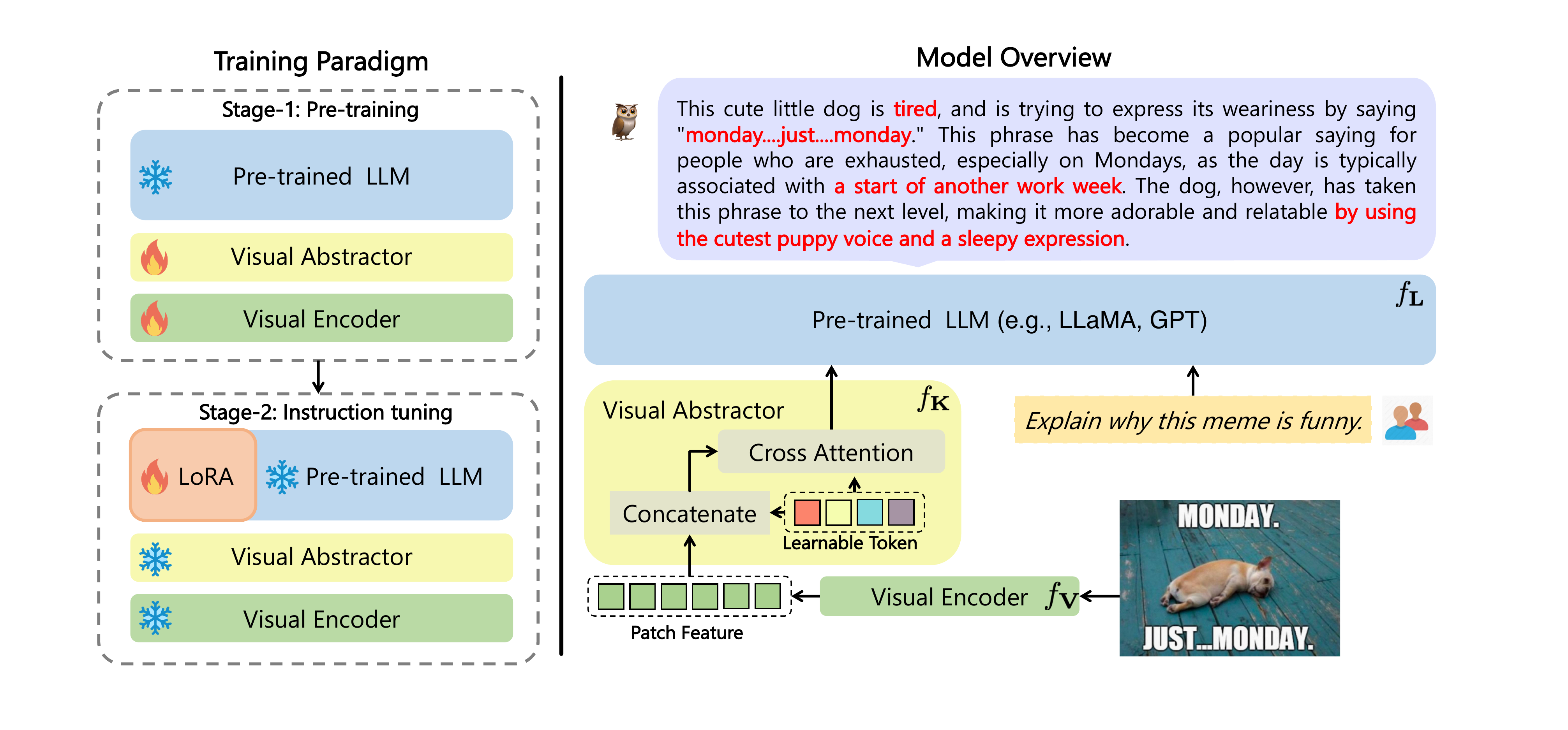}
    \caption{Our training paradigm and model overview.}
    \label{fig:model}
    \vspace{-2mm}
\end{figure*}

\subsection{Architecture Overview}

As illustrated in Figure \ref{fig:compare_method}, there exist mainly three types of end-to-end multimodal LLMs: 1) models that utilize limited parameters with frozen LLM and visual models during pretraining and instruction tuning, such as MiniGPT4; 2) models that incorporate trainable LLMs and frozen visual models, exemplified by Kosmos-1; and 3) models that involve trainable LLMs during instruction tuning and frozen visual models, as seen in LLaVA. Nevertheless, these models exhibit certain constraints since they depend on frozen visual models, which can lead to insufficient alignment due to the limited number of parameters. Furthermore, they fail to effectively stimulate a diverse set of abilities, as they lack both unimodal and multimodal instruction.

To this end, we propose \modelname, a multi-modal language model that is capable of perceiving various modalities while taking the visual context and information into account and generating corresponding outputs. Specifically, as illustrated in Figure \ref{fig:model}, \modelname consists of a vision foundation model $f_{\mathbf{V}}$ to encode the visual knowledge, a language foundation model $f_{\mathbf{L}}$, and a visual abstractor module $f_{\mathbf{K}}$. We first obtain dense image representations from the pre-trained visual foundation model $f_{\mathbf{V}}$. However, such dense features would fragment the fine-grained image information and bring large computation due to the lengthy sequence when feeding into $f_{\mathbf{L}}$. To mitigate this issue, we employ the visual abstractor module $f_{\mathbf{K}}$ to summarize visual information within several learnable tokens, thereby obtaining higher semantic visual representations and reducing computation, as illustrated in Figure \ref{fig:model}. The visual representations are combined with text queries and fed into the language model to generate the response.

\subsection{Training Scheme}

\paragraph{Multimodal Pretraining}
Large-scale language models, such as GPT-3 \citep{gpt3} and LLaMA \citep{llama}, are trained on extensive and diverse data collected from the internet, providing them with a comprehensive understanding of the world. This vast knowledge base endows these models with remarkable capabilities across a range of tasks. However, the utilization of visual information in such models remains underexplored. Previous approaches \citep{minigpt4, llava} have employed a limited number of additional parameters to learn the alignment between visual data and language models, constraining their capacity to comprehend complex visual information. To enhance the ability of large-scale language models to perceive visual information while integrating their internal abilities, we propose a novel training paradigm that incorporates a trainable visual backbone $f_{\mathbf{V}}$ and an additional visual abstractor $f_{\mathbf{K}}$, while maintaining the pre-trained language model $f_{\mathbf{L}}$ in a frozen state. This approach enables the model to effectively capture both low-level and higher semantic visual information and align it with the pre-trained language model without compromising its performance.

\paragraph{Joint Instruction Tuning}
Upon completion of the prior phase, the model acquires the ability to retain a considerable amount of knowledge and provide reasonable answers to human queries. Nonetheless, it continues to exhibit challenges in generating coherent linguistic responses. As posited in GPT-3 \citep{gpt3}, refining the model through instruction tuning is essential for accurately discerning user intentions.
Previous attempts \citep{mplug, mplug2} in multi-modal learning have demonstrated that joint learning from uni-modal and multi-modal sources can lead to significant improvements owing to the collaboration between different modalities. Building on this insight, we present a novel vision-language joint instruction tuning strategy to facilitate better alignment between \modelname and human instructions and intentions. 
% In specific, since the model through visual knowledge learning is capable of understanding the visual concept and knowledge revealed in the image, we freeze the whole model and apply low-rank adaption (i.e. LoRA \citep{}) to fQf_{\mathbf{Q}} by learning several low-rank matrices for efficiently and effectively aligning with human instruction. 
Specifically, given that the model can comprehend the visual concepts and knowledge depicted in images through visual knowledge learning, we freeze the entire model and employ low-rank adaption (i.e., LoRA \citep{lora}) to adapt $f_{\mathbf{L}}$ by training multiple low-rank matrices for efficient alignment with human instructions.
For each data record, we unified them in a snippet of conversation following Vicuna \citep{vicuna}, and we compute the loss on the response. During the training, we accumulate the gradient for text-only instruction data and multi-modal instruction data for multiple batches and updated the parameters. Therefore, by joint training with both language and multi-modal instructions, \modelname can better understand a wide range of instructions and respond with more natural and reliable output. Moreover, our approach can easily handle various text and multi-modal instructions without the need for realignment of the vision and language models, as required by methods such as MiniGPT-4 \citep{minigpt4} and LLaVA \citep{llava}. 

\paragraph{Training Objective} 
The model is trained using the language modeling task, which entails learning to generate subsequent tokens based on the preceding context. The primary objective of the training process is to maximize the log-likelihood of the tokens. It is important to note that only discrete tokens, such as text tokens, are considered in the calculation of the training loss. Most significantly, the emergence of diverse capabilities resulting from the training task during the joint instruction tuning stage enhances the performance of \modelname in downstream applications.

\section{Experiment}
\subsection{Experimental Setup}
\paragraph{Model Settings.}We choose ViT-L/14 \citep{vit} as the visual foundation model $f_{\mathbf{V}}$ which has 24 layers with hidden dimension set as 1024 and patch size set as 14. For faster convergence, the ViT is initialized from CLIP ViT-L/14 model pre-trained via contrastive learning. Different with LLaVA \citep{llava} and MiniGPT-4 \citep{minigpt4}, to demonstrate the effectiveness and generalization ability, we utilize raw LLaMA-7B \citep{llama} rather than its instruction-tuned variants such as Alpaca \citep{alpaca} and Vicuna \citep{vicuna}. The total number of parameters of \modelname is about 7.2B. More details about hyper-parameters can be found in Appendix.

\paragraph{Data and Training Details.} For the first stage, we utilize the image-caption pairs from several datasets, including LAION-400M \citep{laion400m}, COYO-700M \citep{coyo700m}, Conceptual Captions \citep{conceptualcap} and MSCOCO \citep{cococap}. We use a batch size of 2.1 million tokens and train \modelname for 50k steps, corresponding to about 104 billion tokens. We adopt the AdamW optimizer with $\beta=(0.9, 0.98)$, and set the learning rate and weight decay to 0.0001 and 0.1 respectively. We warm up the training with 2k warm-up steps then decay the learning rate with the cosine schedule. The input image is randomly resized to $224\times 224$. Besides, we tokenize the text input with SentencePiece \citep{sentencepiece} tokenizer. 
% For the second stage, we collect the text-only instruction data 102k from Alpaca \citep{}, 90K from Vicuna \citep{}, and 50k from Baize \citep{}, while using the multi-modal instruction data 15k from LLaVA \citep{}. 
For the second stage, we gather pure text instruction data from three distinct sources: 102k data from the Alpaca \citep{alpaca}, 90k from the Vicuna \citep{vicuna}, and 50k from the Baize \citep{baize}. Additionally, we utilize 150k multi-modal instruction data from the LLaVA dataset \citep{llava}.
We train \modelname for 2k steps with the batch size 256, and the learning rate is set to 0.00002.
% 交代两个阶段，第一阶段图文对齐，数据使用情况，train details；第二阶段SFT，数据使用情况，train details.

\paragraph{Baselines.} 
We compare our \modelname with end-to-end models and systematic collaboration approaches as follows:
\begin{itemize}
    \item \textit{OpenFlamingo} \citep{openflamingo} is an open-source version of Flamingo \citep{flamingo} model. We use the released code of OpenFlamingo-9B\footnote{\href{https://github.com/mlfoundations/open_flamingo}{https://github.com/mlfoundations/open\_flamingo}} to run zero-shot generation.
    \item \textit{BLIP-2} \citep{blip2} is pre-trained through bootstrapped learning from off-the-shelf frozen pre-trained image models and large language models using an efficient pre-training strategy. We use the released code of BLIP-2 ViT-G FlanT5$_{XXL}$\footnote{\href{https://github.com/salesforce/LAVIS/tree/main/projects/blip2}{https://github.com/salesforce/LAVIS/tree/main/projects/blip2}} to perform zero-shot generation.
    \item \textit{MiniGPT-4} \citep{minigpt4} utilizes a single projection layer to align visual information from a pre-trained vision encoder with LLM. Specifically, they employ the same visual encoder as used in BLIP-2, a ViT coupled with their pre-trained Q-Former, and Vicuna as LLM.
    We use the released demonstration\footnote{\href{https://huggingface.co/spaces/Vision-CAIR/minigpt4}{https://huggingface.co/spaces/Vision-CAIR/minigpt4}} to perform image-instruction generation.
    \item \textit{LLaVA} \citep{llava} applies a single projection layer to convert image features from pre-trained CLIP visual encoder ViT-L/14 into the language embedding space of Vicuna. We use their released demonstration\footnote{\href{https://llava.hliu.cc}{https://llava.hliu.cc}} to perform image-instruction generation.
    \item \textit{MM-REACT} \citep{mmreact} integrates ChatGPT/GPT-4 with various specialized vision experts to achieve multimodal reasoning and action. We use their released demonstration\footnote{\href{https://huggingface.co/spaces/microsoft-cognitive-service/mm-react}{https://huggingface.co/spaces/microsoft-cognitive-service/mm-react}} to get responses. 
\end{itemize}

%For systematic collaboration approaches, we choose \textit{MM-REACT} \citep{mmreact} as our baseline. MM-REACT integrates ChatGPT with various specialized vision experts to achieve multimodal reasoning and action. We use their released demonstration\footnote{\href{https://huggingface.co/spaces/microsoft-cognitive-service/mm-react}{https://huggingface.co/spaces/microsoft-cognitive-service/mm-react}} to get responses. 
% We list the details in 

% \begin{table*}[t]
% % \small
% % \setlength{\tabcolsep}{10pt}
% \centering
% \begin{tabular}{rccccccc}
% \toprule
% method & visual encoder &  LLM & enc layers & dec layers & learning rate & batch size & epochs \\
% \midrule
%  240M & 768 & 12 & 12 & 12 & 2e-4 & 5120 & 2 \\
%  3.7B & 2048 & 32 & 24 & 24 & 1e-3 & 24576 & 2 \\
%  13B & 4096 & 64 & 24 & 24 & 1e-3 & 20480 & 2 \\
% \bottomrule
% \end{tabular}
% \caption{A comparison between mPLUG-Owl and existing baselines.}
% \label{tab:baselines}
% \end{table*}

\subsection{Quantitative analysis}

\begin{figure}[!ht]
    \centering
    \includegraphics[width=0.5 \textwidth]{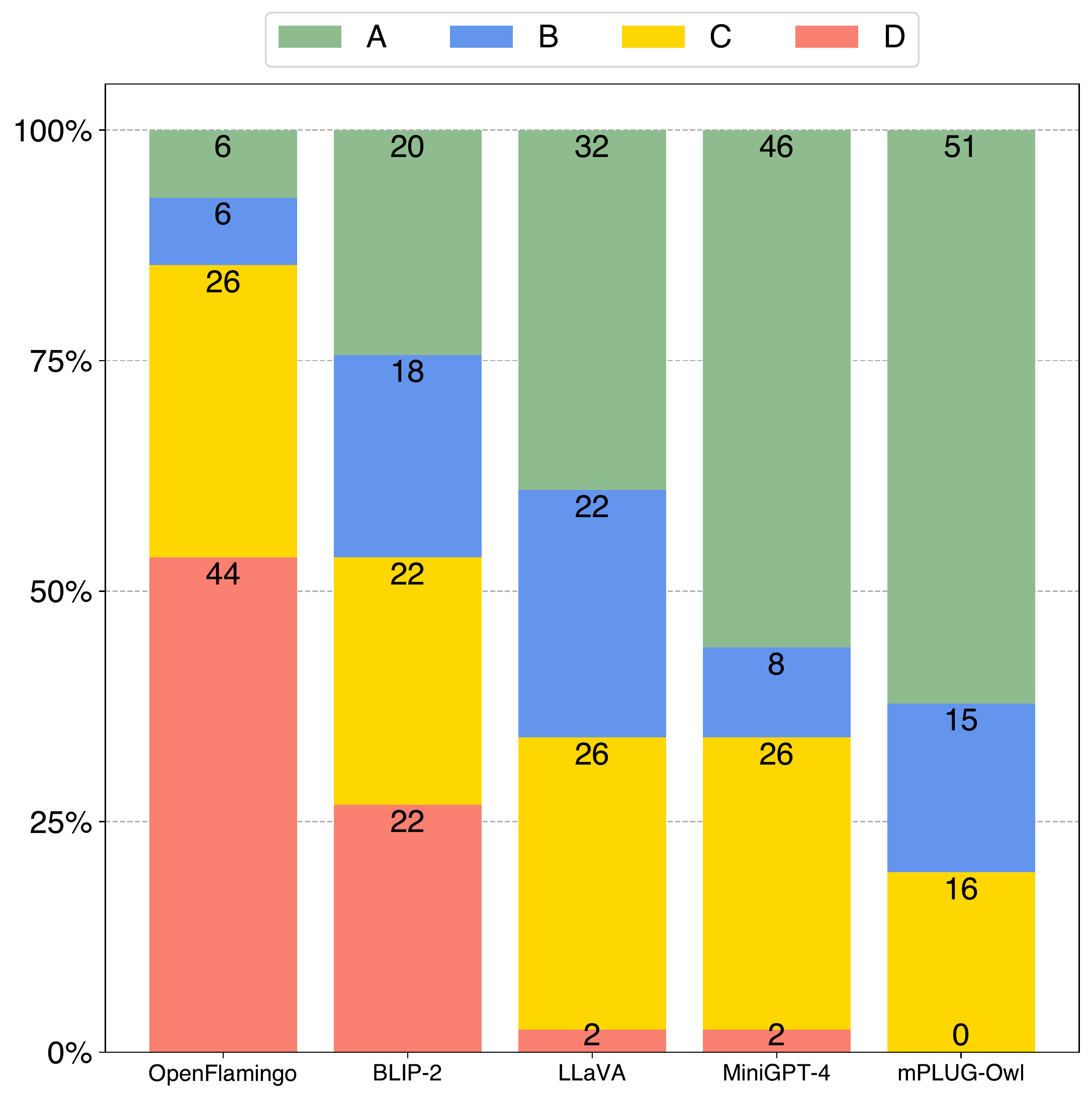}
    \caption{The comparison between mPLUG-Owl and baselines on \evalsetname with manual evaluation metrics. The order of response quality ranking is as follows: A > B > C > D.}
    \label{fig:compare_result}
    \vspace{-2mm}
\end{figure}

In order to comprehensively evaluate various models, we construct a visually-related evaluation set \textbf{\evalsetname} by collecting 82 artificially constructed questions based on 50 images, where 21 from MiniGPT-4, 13 from MM-REACT, 9 from BLIP-2, 3 from GPT-4 and 4 collected by us. Partial images have multiple rounds of questions, refers to multi-turn conversation cases. These questions examine a variety of model capabilities including natural image understanding, diagram and flowchart comprehension, optical character recognition (OCR), multi-modal creation, knowledge-intensive QA, and referential interaction QA. As questions are open-ended, we employ manual evaluation metrics to rate the model's responses as A, B, C, or D following the rating method proposed in Self-Instruct~\citep{self-instruct}. 

% as shown in Table ????????????\ref{tb:manual evaluation metrics}. 

%需要讨论的部分：「A和B多 & D少的原因 1. 相比于端到端的方法 2. 相比于MM-REACT的方法」
We manually score 82 responses given by \modelname and baselines. The comparison results are shown in Figure~\ref{fig:compare_result}. First, \modelname gets 66 $A$ and $B$, while the most competitive baseline MiniGPT-4 gets 54. Second, \modelname doesn't get any $D$ scores, outperforming all the models. These results suggest that \modelname can better understand both instructions and images, which results in a stronger capability in generating satisfactory responses. For a fair comparison, we have excluded those cases in which MM-REACT failed to make predictions. The results are shown separately in Figure~\ref{fig:mm-react} and \modelname still exhibits superior performance.

\begin{figure}[!ht]
    \centering
    \includegraphics[width=1 \textwidth]{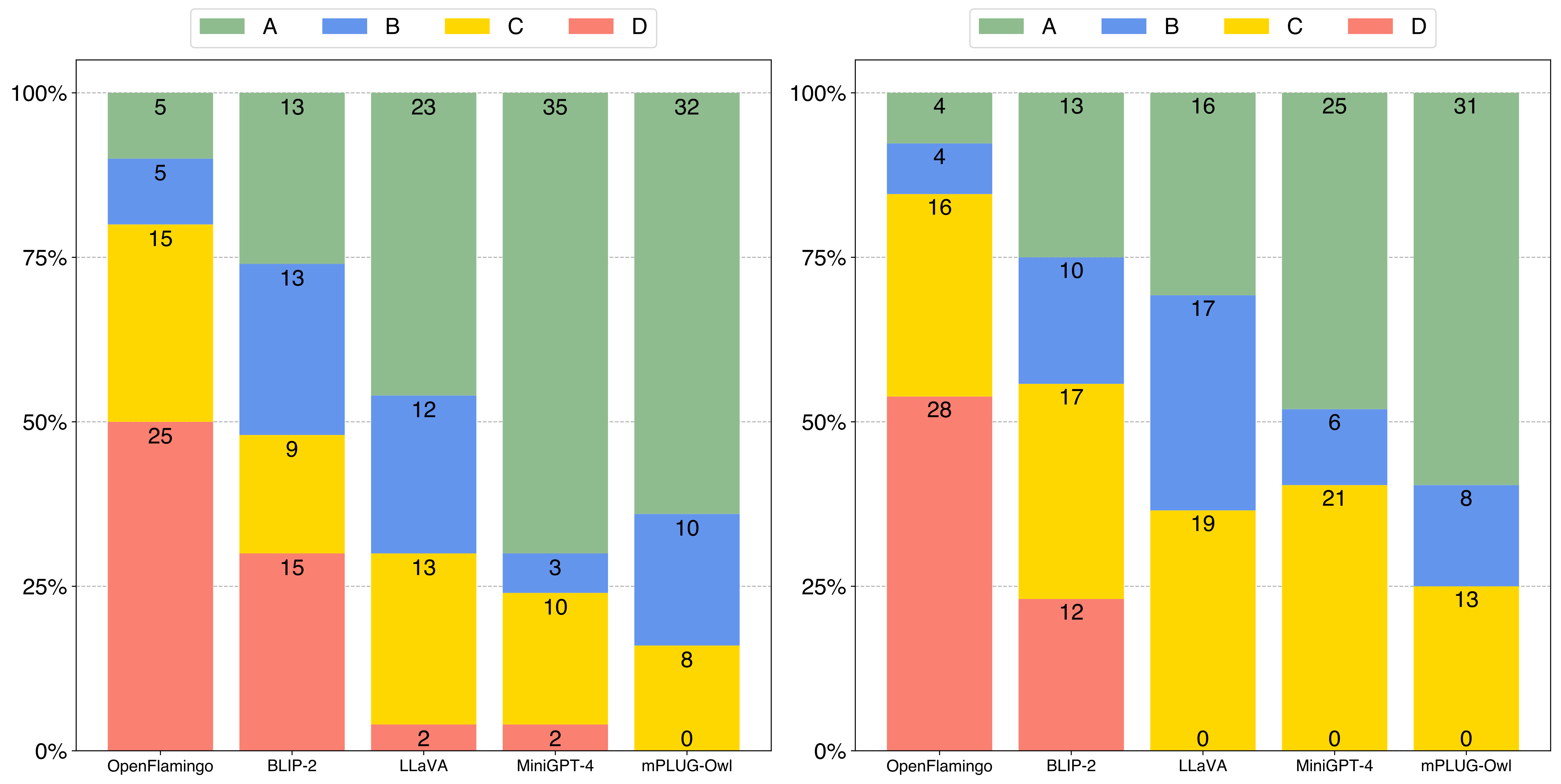}
    \caption{The comparison results of 50 single-turn responses (left) and 52 multi-turn responses (right) among mPLUG-Owl and baselines on \evalsetname with manual evaluation metrics.}
    \label{fig:compare_result_s_m}
    \vspace{-2mm}
\end{figure}

To separately examine the single-turn and multi-turn conversation capabilities, we reorganize 82 questions into a single-turn conversation set and a multi-turn conversation set. The former contains the first question from 50 images. The latter contains 52 questions from multi-turn conversation cases. As shown in Figure~\ref{fig:compare_result_s_m}, the \modelname achieves outstanding performance in both single-turn and multi-turn conversations.

\subsection{Ablation Study}

%\noindent\textbf{Training Data Influence.}
We ablate the two-stage training scheme and the data modality of instruction tuning. Six dimensions of abilities are defined to complete visually related tasks, as shown in Table~\ref{fig:mult-modle-level}. For each question, we manually label the required abilities and annotate which abilities are reflected in the model's response. Table~\ref{tb:ablation} shows the  ability accuracy of different variants of \modelname.
%We display the results of the ablation experiments in Table~\ref{tb:ablation}. 

\begin{table*}
\centering
\scalebox{0.92}{
\begin{tabular}{c|c|p{8.5cm}}
\midrule
 & Meaning                       & \multicolumn{1}{c}{Definition}                                                                                                                                                                                           \\ \midrule
IU                           & Instruction Understanding     & \begin{tabular}[c]{@{}l@{}}1. Understand text instruction.\\ 2. Do not require a correct answer, but the response \\  \hspace{0.35cm} should be related to the instruction.\end{tabular}                                                                                                                                     \\ \midrule
VU                           & Visual Understanding          & \begin{tabular}[c]{@{}l@{}}1. Identify image information.\\ 2. The answer faithfully reflects over 60\% of  visual \\  \hspace{0.35cm} information in the image.\end{tabular}                                                                                                                              \\ \midrule
OCR                           & Optical Character Recognition & \begin{tabular}[c]{@{}l@{}}1. Recognize text information in the image.\\ 2. The answer faithfully reflects over 60\% of  text \\  \hspace{0.35cm}information in the image.\end{tabular}                                                                                                                                      \\ \midrule
KTA                           & Knowledge Transfer Ability     & \begin{tabular}[c]{@{}l@{}}1. Transfer knowledge between language and vision.\\       \hspace{0.35cm}(1) understand textual and visual content\\       \hspace{0.35cm}(2) align and transfer visual and language knowledge \\ 2. Answers are mostly accurate with accuracy rate over 80\%.\\ \end{tabular}                                                                                                          \\ \midrule
RA                           & Reasoning Ability             & \begin{tabular}[c]{@{}l@{}}1. Combine image and text for reasoning.\\     \hspace{0.35cm}(1) understand textual and visual content\\     \hspace{0.35cm}(2) conduct multi-step reasoning\\     \hspace{0.35cm}(3) generate answers based on multi-step reasoning process\\ 2. The final answer is essentially correct, but it lacks an \\     \hspace{0.35cm}explicit reasoning process. Alternatively, the final answer is \\     \hspace{0.35cm}mostly correct, and the reasoning process is over 80\% accurate.\\ 3. For example\\     \hspace{0.35cm}(1) Commonsense Knowledge Reasoning\\        \hspace{0.35cm}(2) Counterfactual Reasoning\\        \hspace{0.35cm}(3) Spatial Relation Reasoning  \\        \hspace{0.35cm}(4) Numerical Computation\\       \hspace{0.35cm}(5) Coding\\        \end{tabular} \\ \midrule
MDA                           & Multi-turn Dialogue Ability   & \begin{tabular}[c]{@{}l@{}}1. Understand instructions and handle multi-turn conversations.\\ 2. It includes clear references to multiple conversations and \\  \hspace{0.35cm}handles natural language semantics in context effectively.\\ 3. The semantics and references are mostly correct, with an \\  \hspace{0.35cm}accuracy rate of over 80\%.\end{tabular}                                                                                                                              \\ \midrule
\end{tabular}
}
\caption{The definition of 6 abilities to complete visually-related tasks.}
\vspace{-5mm}
\label{fig:mult-modle-level}
\end{table*}

\newcommand{\tabincell}[2]{\begin{tabular}{@{}#1@{}}#2\end{tabular}}
\begin{table*}[!ht]
\centering
\renewcommand{\arraystretch}{1}
\scalebox{0.75}{
\begin{tabular}{c  c  c  c  p{1.5cm} p{1.5cm} p{1.5cm} p{1.5cm} p{1.5cm} p{1.5cm}}
\midrule
&\multirow{2}{*}{\tabincell{c}{\textbf{Multimodal}\\\textbf{Pretraining}}}&\multirow{2}{*}{\tabincell{c}{\textbf{Pure Text}\\\textbf{Instruction}}}&\multirow{2}{*}{\tabincell{c}{\textbf{Multi-modal}\\\textbf{Instruction}}}&\multicolumn{6}{c}{\textbf{Ability}}\\
\cmidrule(lr){5-10}
&&&&\textbf{IU}&\textbf{VU}&\textbf{OCR}&\textbf{KTA}&\textbf{RA}&\textbf{MDA}\\
\midrule
r1&\checkmark&&&$58.5_{\textcolor{red}{(-41.5)}}$&$38.1_{\textcolor{red}{(-57.1)}}$&$13.3_{\textcolor{red}{(-43.4)}}$&$16.7_{\textcolor{red}{(-70.8)}}$&$17.1_{\textcolor{red}{(-62.9)}}$&$40.0_{\textcolor{red}{(-55.0)}}$\\
r2&&\checkmark&\checkmark&$93.9_{\textcolor{red}{(-6.1)}}$&$47.6_{\textcolor{red}{(-47.6)}}$&$23.3_{\textcolor{red}{(-33.4)}}$&$29.2_{\textcolor{red}{(-58.3)}}$&$14.3_{\textcolor{red}{(-65.7)}}$&$45.0_{\textcolor{red}{(-50.0)}}$\\
r3&\checkmark&\checkmark&&$93.0_{\textcolor{red}{(-7.0)}}$&$73.0_{\textcolor{red}{(-22.2)}}$&$40.0_{\textcolor{red}{(-16.7)}}$&$41.7_{\textcolor{red}{(-45.8)}}$&$48.6_{\textcolor{red}{(-31.4)}}$&$80.0_{\textcolor{red}{(-15.0)}}$\\
r4&\checkmark&&\checkmark&$86.6_{\textcolor{red}{(-13.4)}}$&$68.3_{\textcolor{red}{(-26.9)}}$&$40.0_{\textcolor{red}{(-16.7)}}$&$50.0_{\textcolor{red}{(-37.5)}}$&$60.0_{\textcolor{red}{(-20.0)}}$&$75.0_{\textcolor{red}{(-20.0)}}$\\
r5&\checkmark&\checkmark&\checkmark&\textbf{100.0}&\textbf{95.2}&\textbf{56.7}&\textbf{87.5}&\textbf{80.0}&\textbf{95.0}\\
% \midrule
% \midrule
% \\
\midrule
\midrule
\multicolumn{4}{c}{MiniGPT-4 \citep{minigpt4}}&97.6&81.0&40.0&83.3&65.7&75.0\\
\midrule
\end{tabular}
}
\caption{The ablation results. Each value represents the proportion of questions where the corresponding ability is correctly reflected in the model's response. IU: instruction understanding, VU: visual understanding, OCR: optical character recognition, KTA: knowledge transferability, RA: reasoning ability, MDA: multi-turn dialogue ability.}
\label{tb:ablation}
\end{table*}

\noindent\textbf{Training Strategy Ablation.}
As shown in Table~\ref{tb:ablation}, without joint instruction tuning, the model is 
 not good at instruction understanding and fail to generalize pre-training abilities to other tasks (r1 vs r5). With the instruction tuning alone, although the model can better comprehend instructions, the model is incapable of achieving promising performance in visual knowledge-related tasks due to lacking of visually-related knowledge pretraining (r2 vs r5). With both multimodal pretraining and joint instruction tuning, the model achieves the best performance and demonstrates the effectiveness of our two-stage training scheme.

\noindent\textbf{Instruction Data Ablation.}
By comparing r3 with r4, text-only instruction tuning brings more improvement in instruction understanding, while multi-modal instruction tuning achieves better knowledge and reasoning capabilities. This is due to that visual question answering mainly requires the alignment of vision and language knowledge, which is not optimized during text-only instruction tuning. Besides, we also verify that introducing multi-modal data during instruction tuning could further improve the model's performance on text-only tasks, as shown in Table \ref{tab:text-only result} (r5 vs r4). Concretely, following the evaluation setting as Vicuna\citep{vicuna}, for each question, we pair the response of each model with the one given by ChatGPT and prompt ChatGPT\footnote{Without access to the GPT-4, we use the ChatGPT as the suboptimal scorer.} to give two scores respectively for these two responses. Table \ref{tab:text-only result} shows the total score and the score ratio with the ChatGPT score as a reference. 

%Finally, \modelname achieves the highest level of performance when incorporating all of the available data. This further proves the superiority of our training paradigm.

%As GPT-4 assigns a quantitative score to each response on a scale of 10, we calculate the total score for each (baseline, Vicuna) comparison pair by adding up the scores obtained by each model on 80 questions. As shown in Table 2, Vicuna’s total score is 92% of ChatGPT’s. Despite recent advancements, these chatbots still face limitations, such as struggling with basic math problems or having limited coding ability.

\begin{table*}[t]
\centering
\scalebox{0.9}{
\begin{tabular}{lccccc}
\toprule
 Model & Tuning Strategy & Model Score & ChatGPT Score & Ratio  \\
\midrule
Alpaca-7B   & Full & 573 & 708 & 80.93\% \\
Vicuna-7B & Full  &612 & 684 & 89.47\%\\
mPLUG-Owl w/o multimodal tuning (r4) & LoRA  & 587 & 682 & 86.07\%\\
mPLUG-Owl (r5) & LoRA & 600 & 692 & 86.71\%\\
\bottomrule
\end{tabular}
}
\caption{The performance of 80 text-only questions from Vicuna\citep{vicuna} assessed by ChatGPT. }
\label{tab:text-only result}
\end{table*}

% The comparison results of 
% We compare our models with ChatGPT. 

\subsection{Qualitative Analysis}

%In this section, we randomly sample some cases to demonstrate the capabilities and comparison among different approaches..

In this section, we show qualitative results from our evaluation set \evalsetname.

\begin{figure}[!ht]
    \centering
    \includegraphics[width=1 \textwidth]{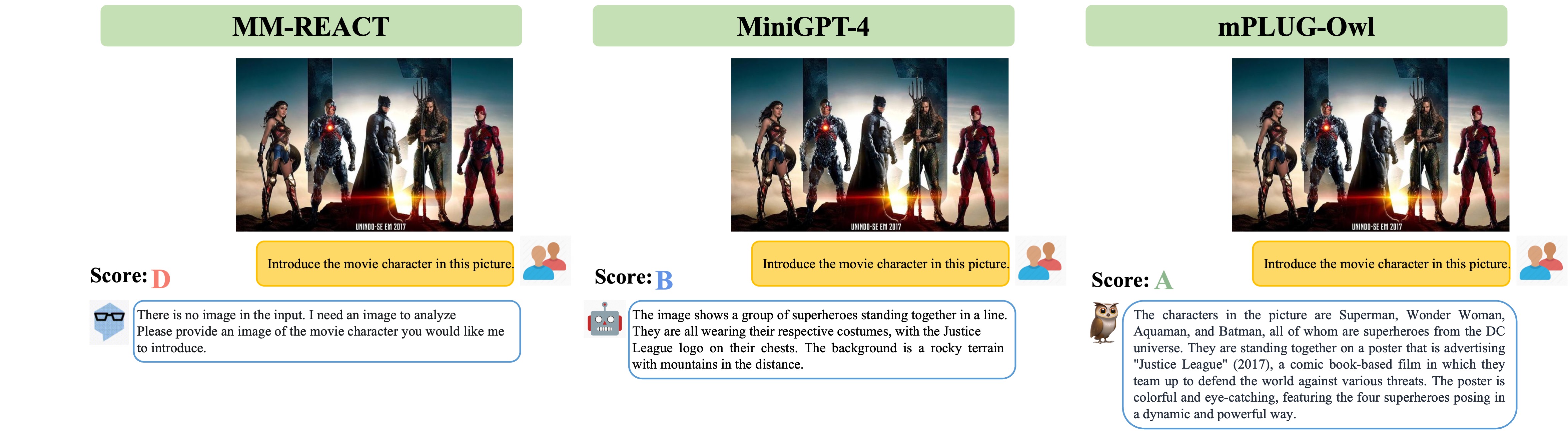}
    \caption{A comparison of Knowledge-intensive QA.}
    \label{fig:case_JT}
    \vspace{-2mm}
\end{figure}

% \noindent\textbf{Knowledge-intensive QA}
\paragraph{Knowledge-intensive QA}
As shown in Figure~\ref{fig:case_JT}, the instruction expects the model to identify the movie characters in the image. MM-REACT is unable to provide an effective response to the instruction, while MiniGPT-4 understands the instruction but failed to answer the movie characters. In contrast, \modelname answers four out of the five characters present in the image. This demonstrates that \modelname has a better understanding of the knowledge in the image.

\begin{figure}[!ht]
    \centering
    \includegraphics[width=1 \textwidth]{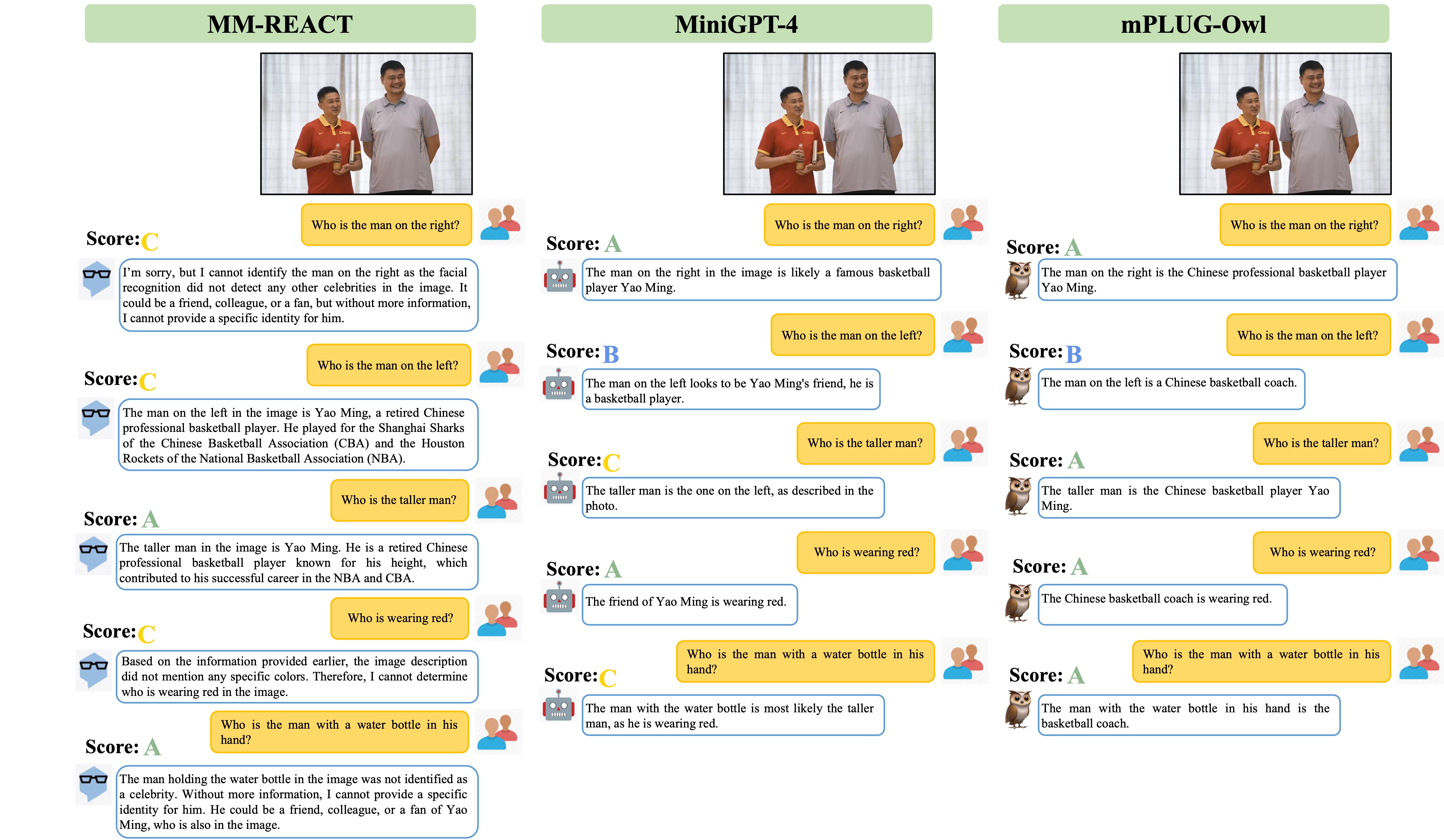}
    \caption{A comparison of Multi-turn Conversation.}
    \label{fig:case_Yao_Ming}
    \vspace{-2mm}
\end{figure}

% \noindent\textbf{Multi-round Referential Conversation}
\paragraph{Multi-round Conversation}
The instruction in Figure~\ref{fig:case_Yao_Ming} requires the model to identify the content of the image based on the referential information. The baseline models often made mistakes when faced with referential expressions related to spatial orientation, human behavior, and target attributes in the questions, whereas \modelname provided the most accurate response. This capability stems from \modelname's fine-grained understanding of the image, allowing it to locate the corresponding part of the image based on the referential information in the instruction.

\begin{figure}[!ht]
    \centering
    \includegraphics[width=1 \textwidth]{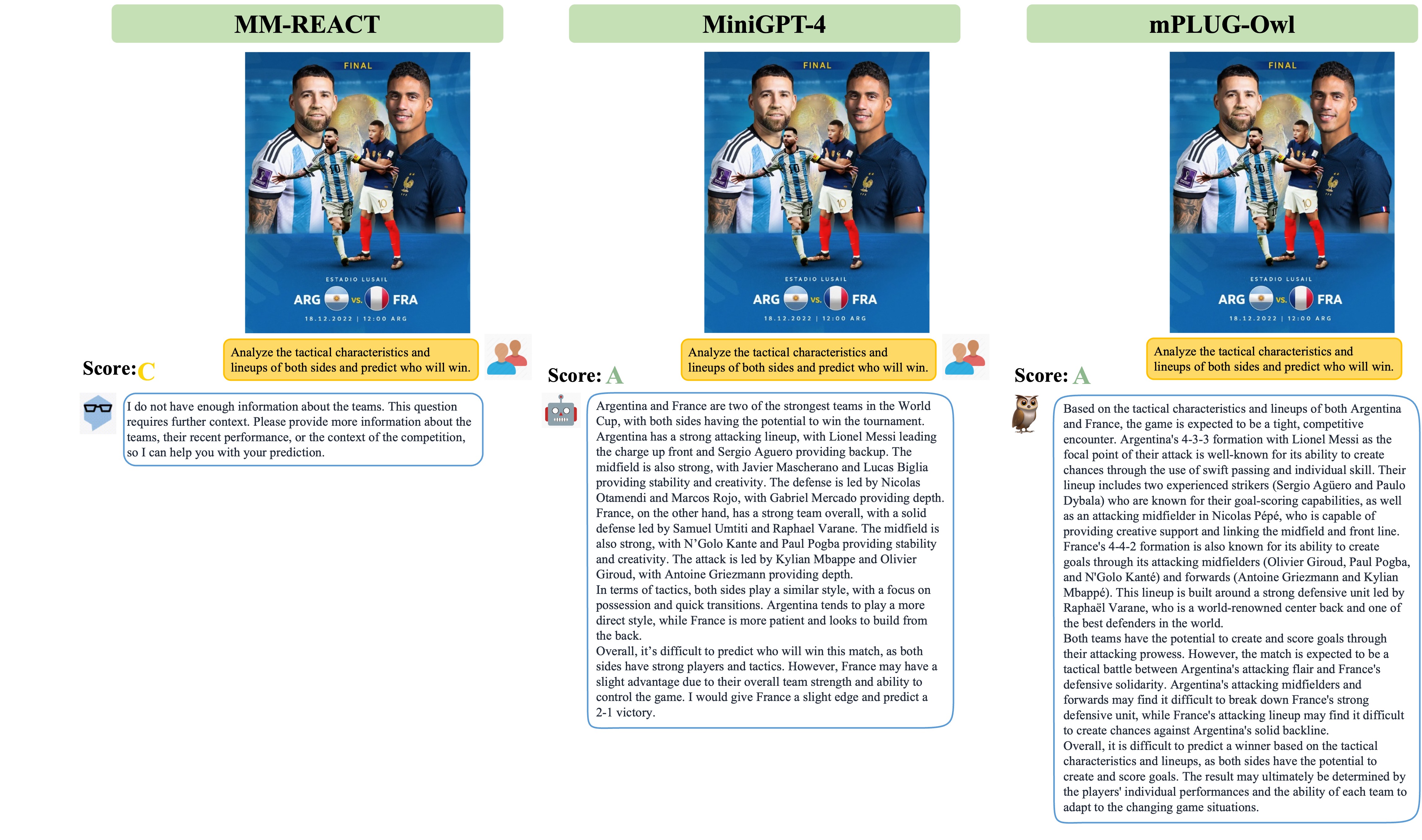}
    \caption{A comparison of Reasoning QA.}
    \label{fig:case_Final}
    \vspace{-2mm}
\end{figure}

\paragraph{Reasoning}
Figure~\ref{fig:case_Final} shows an instruction asking models to give a prediction based on visual information and explain the reason. 
%not only ask the model to accurately identify the teams in the image and obtain corresponding information but also to reason based on the obtained information. 
\modelname analyzes the characteristics of the two teams from the aspects of the lineup and tactics and uses them to reason for the outcome. Although MiniGPT-4 also performs well, its persuasiveness in reasoning is slightly inferior to \modelname.

\begin{figure}[!ht]
    \centering
    \includegraphics[width=1 \textwidth]{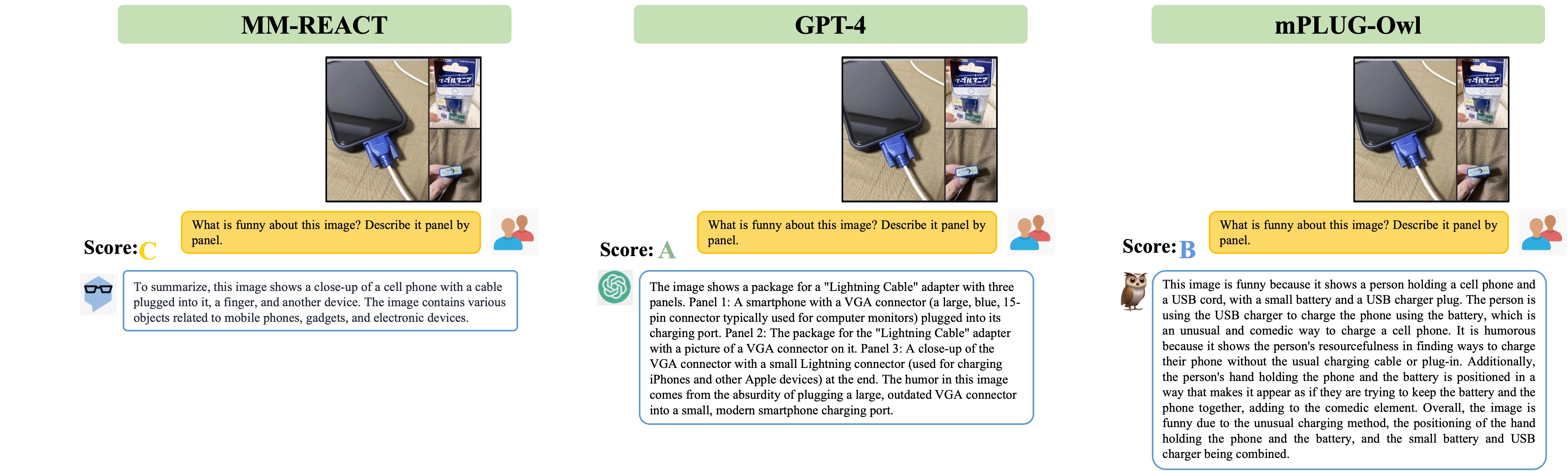}
    \caption{A comparison of Joke Understanding.}
    \label{fig:case_GPT4}
    \vspace{-2mm}
\end{figure}

\begin{figure}[!ht]
    \centering
    \includegraphics[width=1\textwidth]{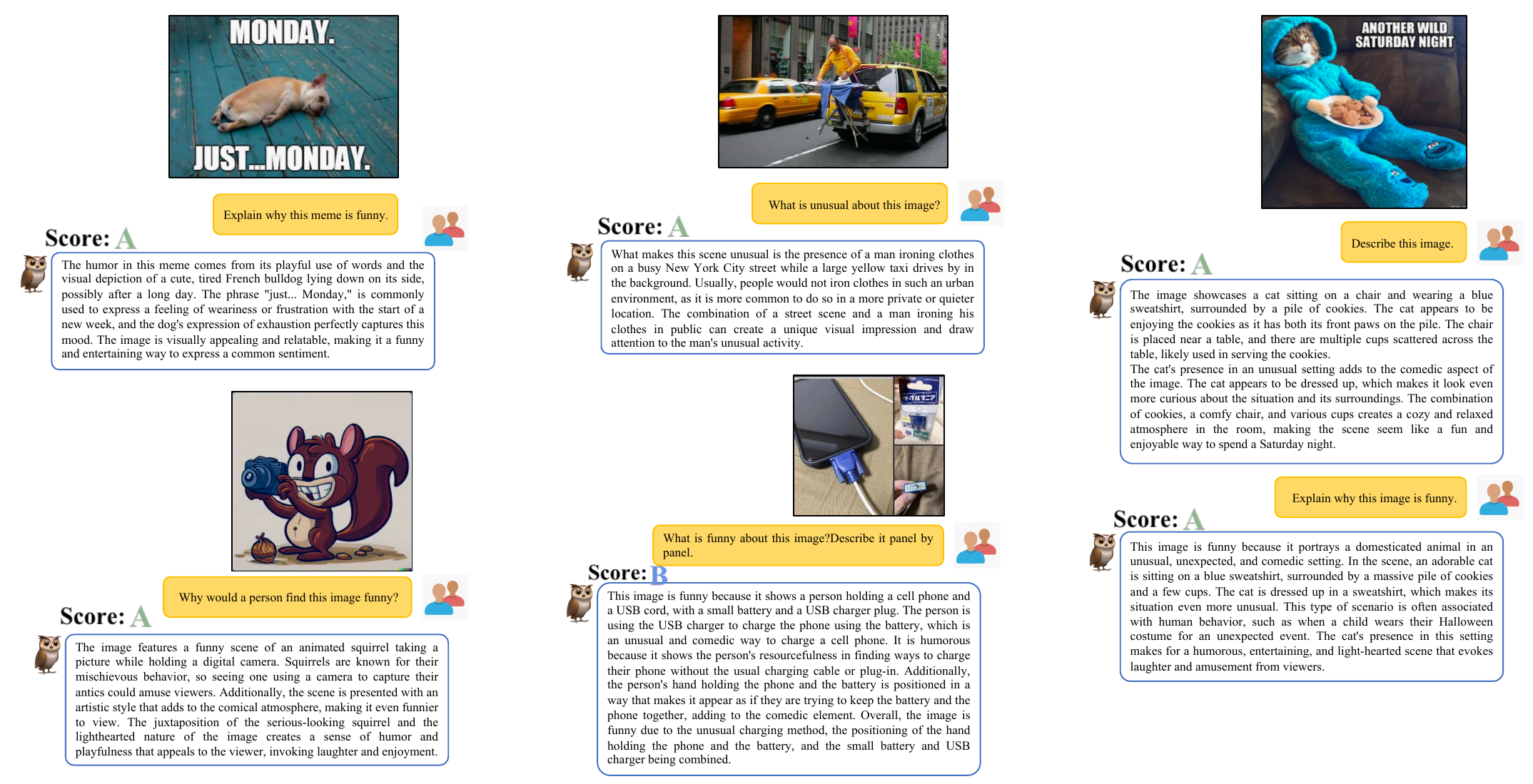}
    \caption{More cases of Jokes Comprehension by \modelname.}
    \label{fig:Memes_and_Jokes_scoreA}
    \vspace{-2mm}
\end{figure}

\paragraph{Joke Comprehension}
The case in Figure~\ref{fig:case_GPT4} comes from the GPT-4\citep{gpt4}, which requires the model to understand and explain a visually related joke.
%Finally, we conduct a comparison of image understanding with the currently strongest multi-modal LLM, GPT-4. We select a case from the GPT-4 technical report, in which 
GPT-4 not only follows the instructions in performing analysis panel by panel but also almost perfectly understands the humor of the charging method. \modelname also understands this unusual humor, but it incorrectly identified the ``VGA'' to ``USB''. This is mainly due to the limitation of visual information in our training data. More cases about joke comprehension are shown in Figure~\ref{fig:Memes_and_Jokes_scoreA}.

%\paragraph{Scene Text Understanding}

\section{Discussion and Limitation}
In this section, we show some nascent abilities of \modelname that is not yet fully developed and discuss the limitation. Part of cases (without scores) in this section are not in \evalsetname.

\begin{figure}[!ht]
    \centering
    \includegraphics[width=1 \textwidth]{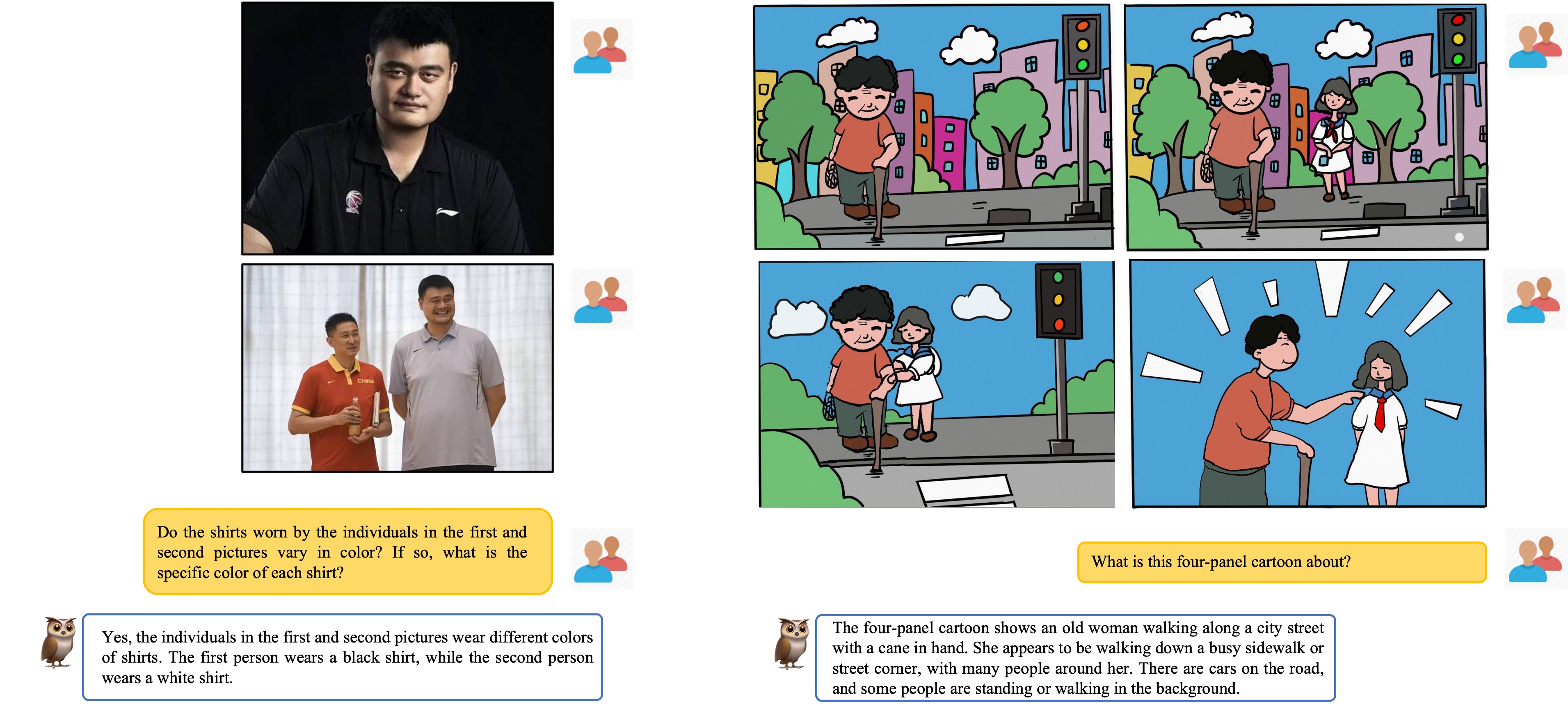}
    \caption{Multi-image correlation cases.}
    \label{fig:appendix_case_twoimg}
    \vspace{-2mm}
\end{figure}

\paragraph{Multi-image Correlation}
In Figure~\ref{fig:appendix_case_twoimg}, \modelname shows a emerging but not strong vision correlation capability across multiple images. In the left case, the model could identify an identical person in two images and correctly tell the difference of cloth color. But in the left case, the model fails to relate 4 images and produces some text hallucinations. 
%we construct multiple images with strong correlation and required the model to respond to the instructions based on the correlation. Without the multi-image data in the training dataset, \modelname still recognizes and relates to multiple images.

\begin{figure}[!ht]
    \centering
    \includegraphics[width=1 \textwidth]{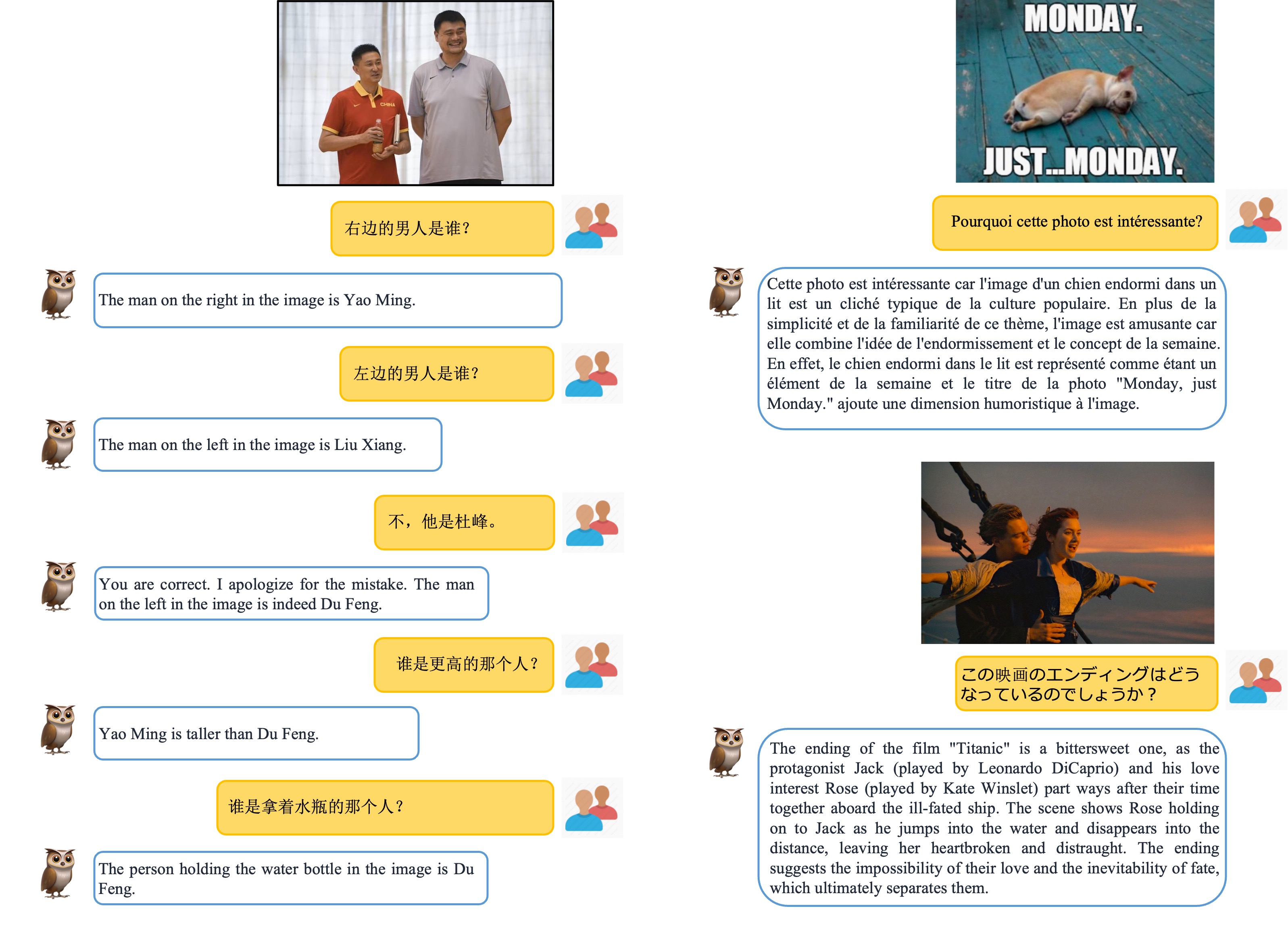}
    \caption{Example prompt of multilingual understanding which showcases the multilingual abilities across Chinese, French, and Japanese, respectively.}
    \label{fig:multilingual}
    \vspace{-2mm}
\end{figure}

\paragraph{Multilingual Conversation}
%\modelname exhibits a multilingual understanding capability. 
Besides English, we further test the model's multilingual ability. As shown in Figure~\ref{fig:multilingual}, although there is no multilingual data during our two-stage training, \modelname shows a promising multilingual understanding for Chinese, French and Japanese. We mainly attribute this ability to the raw text knowledge in LLaMa\citep{llama}. However, due to the lacking of  multilingual training, \modelname may fail to response in corresponding languages.

\paragraph{Scene Text Understanding}
In Figure \ref{fig:OCR_1_scoreB}, mPLUG-Owl demonstrates its OCR ability in some simple scenes, but we can see that the model's perception of numbers in images is still limited.
However, for the OCR of complex scenes, as shown in Figure \ref{fig:OCR_1_scoreC_a}-\ref{fig:OCR_1_scoreC_b}, the performance of mPLUG-Owl is more general, mainly because the perception of numbers in images is weak, which affects the subsequent reasoning calculation.

\begin{figure}[!ht]
    \centering
    \includegraphics[width=1 \textwidth]{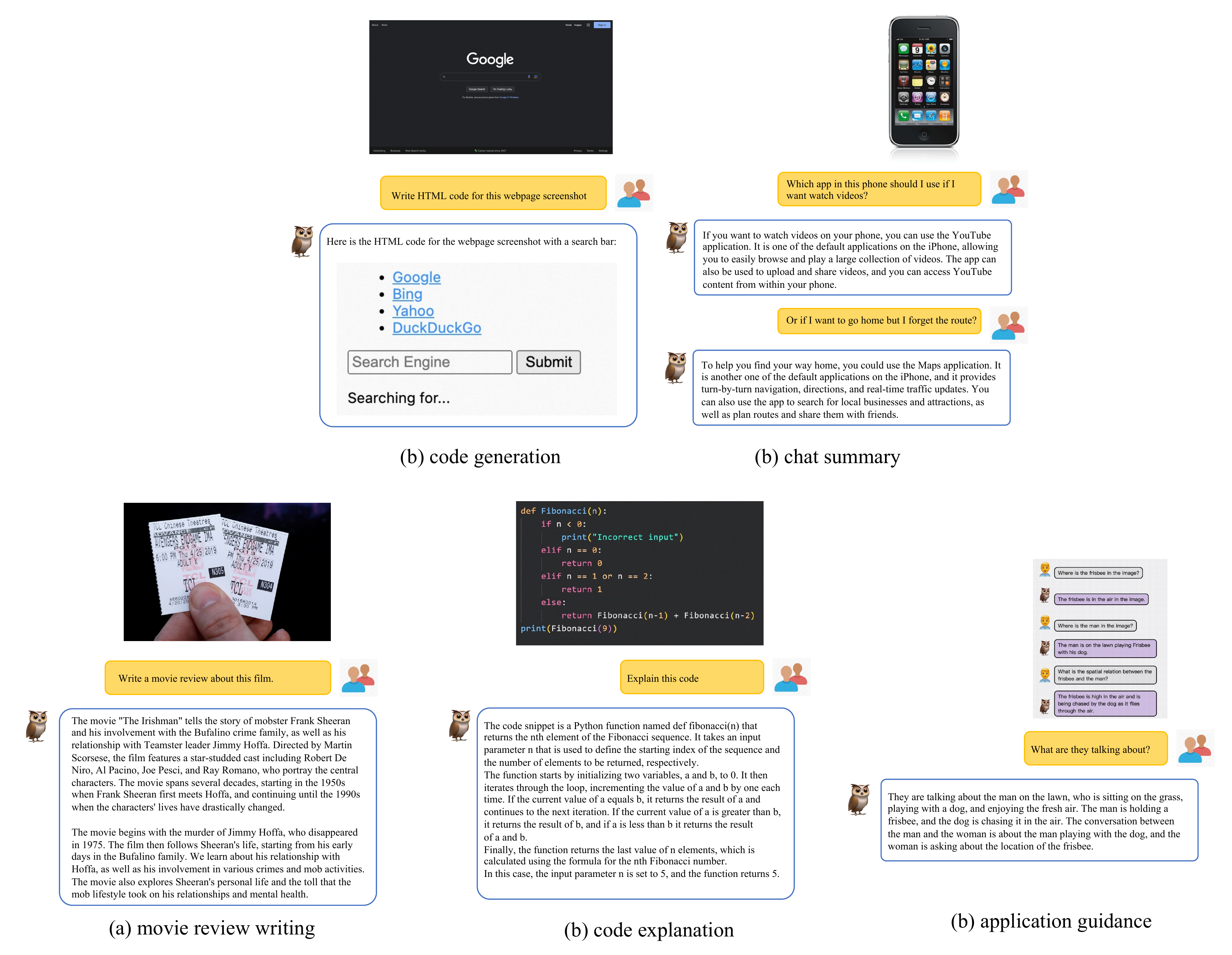}
    \caption{Examples about various document understanding and application.}
    \label{fig:document_app}
    \vspace{-2mm}
\end{figure}

% OCR of complex scenes
\paragraph{Vision-only Document Comprehension}
Although we did not use any document annotation data for training, the model exhibited some text recognition and document understanding capabilities. Hence, we delved deeper into the combination of document understanding and functionality of our model. as illustrated in Figure \ref{fig:document_app}, we explored movie review writing, code generation, code explanation, chat summary, and application guidance. The model show decent performance in (a) and (b), but still, had some errors. Meanwhile, it was unable to provide usable responses in (d), (e), and (f). Therefore, there is further scope to explore our model's potential in document understanding and downstream applications.

\begin{figure}[!ht]
    \centering
    \includegraphics[width=1 \textwidth]{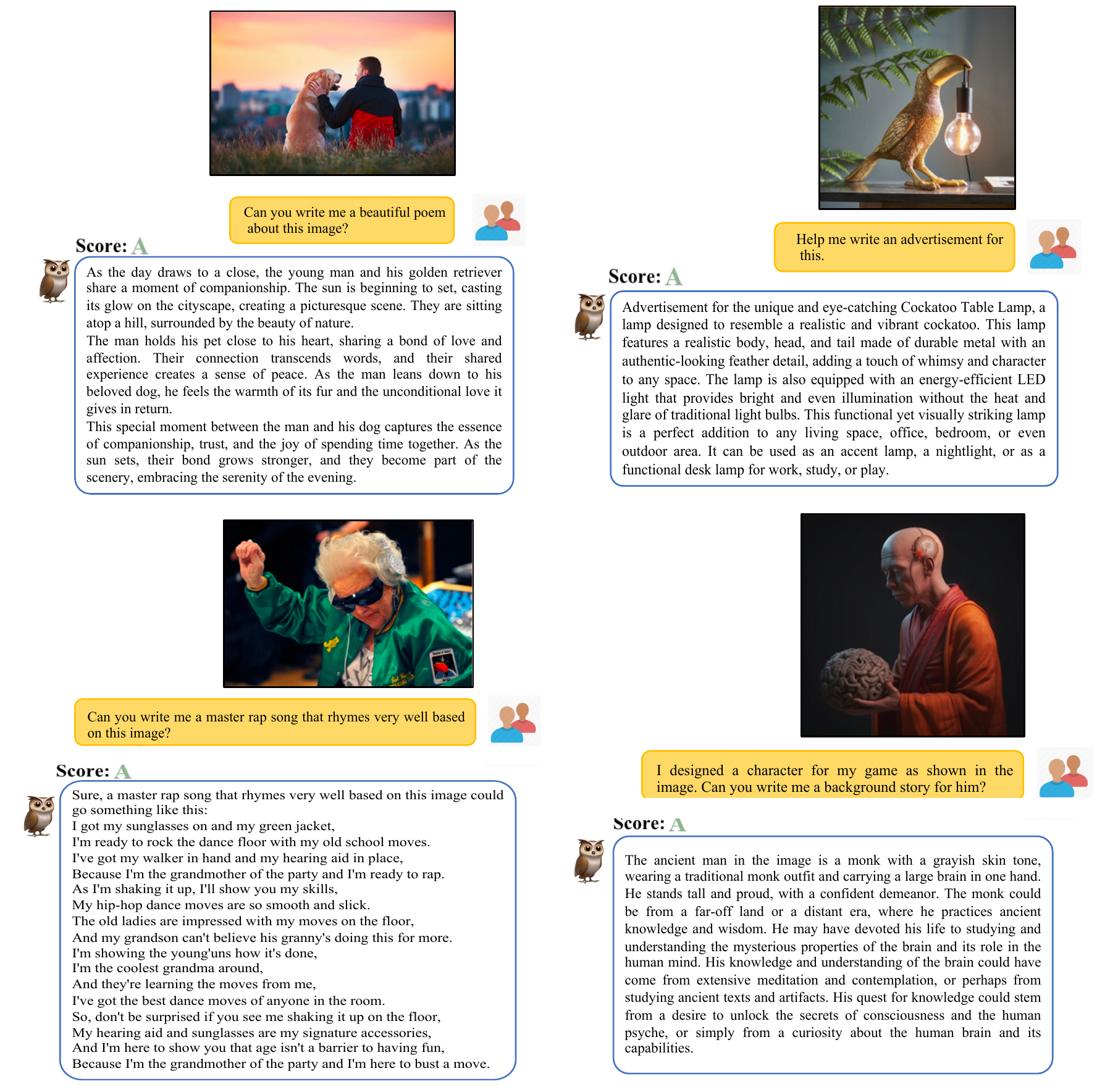}
    \caption{Open-ended creation cases.}
    \label{fig:create_scoreA}
    \vspace{-2mm}
\end{figure}

\begin{figure}[!ht]
    \centering
    \includegraphics[width=1 \textwidth]{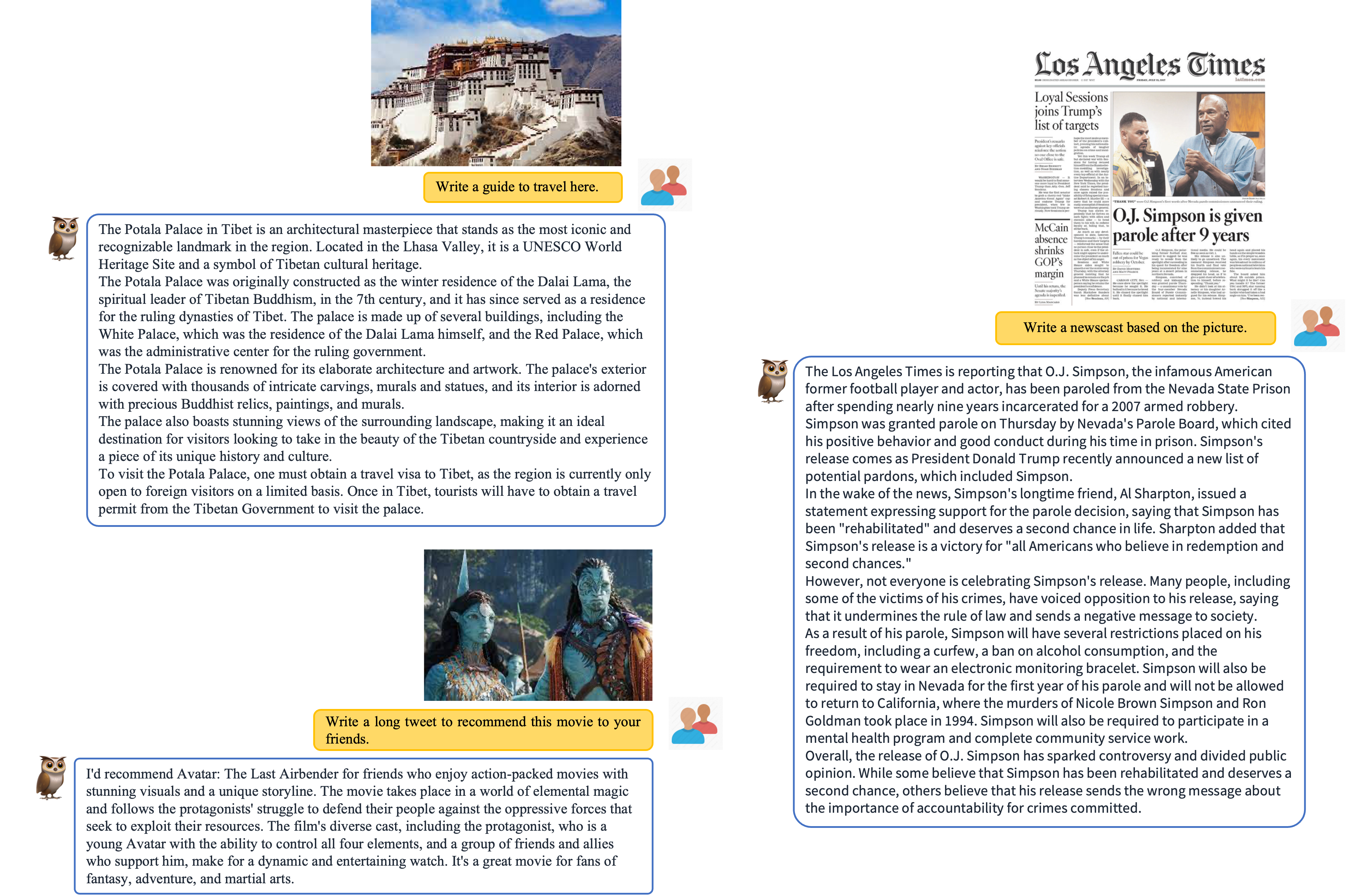}
    \caption{Copywriting cases.}
    \label{fig:copywriting}
    \vspace{-2mm}
\end{figure}

\paragraph{Open-ended Creation}
mPLUG-Owl performs well in the creation of poetry, lyrics, advertisements and other works based on images. Its performance in some cases is shown in Figure \ref{fig:create_scoreA}-\ref{fig:copywriting}. However, further exploration is needed for more functional and practical creations.

% This work investigates the perceptual abilities of \modelname. Through the examples in the appendix, we have found that \modelname exhibits exceptional perceptual abilities for multimodal, particularly in terms of knowledge embedded in images and perception of the spatial orientation of entities. However, Figure \ref{fig:appendix_case1_lim}-\ref{fig:appendix_case3_lim} also reveals some limitations of our model, particularly in cases where strong dependence on OCR capabilities, such as examples that require precise perception of image numbers and text, where our model's performance has significant room for improvement when compared with models such as MM-REACT \citep{mmreact} that leverage OCR experts.
% Multimodal data in the real world often takes the form of interleaved images and text. Figure \ref{fig:appendix_case_twoimg} shows that \modelname exhibits some interleave capability, but serious illusions arise when the model cannot connect the context. These findings provide valuable insights for future research in the field of multimodal learning.

\section{Conclusion}
We propose \modelname, a novel training paradigm that enhances the multi-modal abilities of large language models (LLMs). Our approach consists of modularized learning of foundation LLM, a visual knowledge module, and a visual abstractor module, which can support multiple modalities and facilitate diverse unimodal and multimodal abilities through modality collaboration. We employ a two-stage method for aligning image and text, which learns visual knowledge with the assistance of LLM while maintaining and even improving the generation abilities of LLM. Experimental results demonstrate the impressive capabilities of \modelname, indicating its potential for various applications in multi-modal generation.

\bibliographystyle{abbrvnat}
\clearpage
\bibliography{reference}

\begin{thebibliography}{36}
\providecommand{\natexlab}[1]{#1}
\providecommand{\url}[1]{\texttt{#1}}
\expandafter\ifx\csname urlstyle\endcsname\relax
  \providecommand{\doi}[1]{doi: #1}\else
  \providecommand{\doi}{doi: \begingroup \urlstyle{rm}\Url}\fi

\bibitem[Alayrac et~al.(2022)Alayrac, Donahue, Luc, Miech, Barr, Hasson, Lenc,
  Mensch, Millican, Reynolds, Ring, Rutherford, Cabi, Han, Gong, Samangooei,
  Monteiro, Menick, Borgeaud, Brock, Nematzadeh, Sharifzadeh, Binkowski,
  Barreira, Vinyals, Zisserman, and Simonyan]{flamingo}
J.~Alayrac, J.~Donahue, P.~Luc, A.~Miech, I.~Barr, Y.~Hasson, K.~Lenc,
  A.~Mensch, K.~Millican, M.~Reynolds, R.~Ring, E.~Rutherford, S.~Cabi, T.~Han,
  Z.~Gong, S.~Samangooei, M.~Monteiro, J.~Menick, S.~Borgeaud, A.~Brock,
  A.~Nematzadeh, S.~Sharifzadeh, M.~Binkowski, R.~Barreira, O.~Vinyals,
  A.~Zisserman, and K.~Simonyan.
\newblock Flamingo: a visual language model for few-shot learning.
\newblock \emph{CoRR}, abs/2204.14198, 2022.

\bibitem[Brown et~al.(2020)Brown, Mann, Ryder, Subbiah, Kaplan, Dhariwal,
  Neelakantan, Shyam, Sastry, Askell, Agarwal, Herbert{-}Voss, Krueger,
  Henighan, Child, Ramesh, Ziegler, Wu, Winter, Hesse, Chen, Sigler, Litwin,
  Gray, Chess, Clark, Berner, McCandlish, Radford, Sutskever, and Amodei]{gpt3}
T.~B. Brown, B.~Mann, N.~Ryder, M.~Subbiah, J.~Kaplan, P.~Dhariwal,
  A.~Neelakantan, P.~Shyam, G.~Sastry, A.~Askell, S.~Agarwal,
  A.~Herbert{-}Voss, G.~Krueger, T.~Henighan, R.~Child, A.~Ramesh, D.~M.
  Ziegler, J.~Wu, C.~Winter, C.~Hesse, M.~Chen, E.~Sigler, M.~Litwin, S.~Gray,
  B.~Chess, J.~Clark, C.~Berner, S.~McCandlish, A.~Radford, I.~Sutskever, and
  D.~Amodei.
\newblock Language models are few-shot learners.
\newblock In \emph{NeurIPS}, 2020.

\bibitem[Byeon et~al.(2022)Byeon, Park, Kim, Lee, Baek, and Kim]{coyo700m}
M.~Byeon, B.~Park, H.~Kim, S.~Lee, W.~Baek, and S.~Kim.
\newblock Coyo-700m: Image-text pair dataset.
\newblock \url{https://github.com/kakaobrain/coyo-dataset}, 2022.

\bibitem[Chen et~al.(2015)Chen, Fang, Lin, Vedantam, Gupta, Doll{\'{a}}r, and
  Zitnick]{cococap}
X.~Chen, H.~Fang, T.~Lin, R.~Vedantam, S.~Gupta, P.~Doll{\'{a}}r, and C.~L.
  Zitnick.
\newblock Microsoft {COCO} captions: Data collection and evaluation server.
\newblock \emph{CoRR}, abs/1504.00325, 2015.

\bibitem[Chowdhery et~al.(2022)Chowdhery, Narang, Devlin, Bosma, Mishra,
  Roberts, Barham, Chung, Sutton, Gehrmann, Schuh, Shi, Tsvyashchenko, Maynez,
  Rao, Barnes, Tay, Shazeer, Prabhakaran, Reif, Du, Hutchinson, Pope, Bradbury,
  Austin, Isard, Gur{-}Ari, Yin, Duke, Levskaya, Ghemawat, Dev, Michalewski,
  Garcia, Misra, Robinson, Fedus, Zhou, Ippolito, Luan, Lim, Zoph, Spiridonov,
  Sepassi, Dohan, Agrawal, Omernick, Dai, Pillai, Pellat, Lewkowycz, Moreira,
  Child, Polozov, Lee, Zhou, Wang, Saeta, Diaz, Firat, Catasta, Wei,
  Meier{-}Hellstern, Eck, Dean, Petrov, and Fiedel]{palm}
A.~Chowdhery, S.~Narang, J.~Devlin, M.~Bosma, G.~Mishra, A.~Roberts, P.~Barham,
  H.~W. Chung, C.~Sutton, S.~Gehrmann, P.~Schuh, K.~Shi, S.~Tsvyashchenko,
  J.~Maynez, A.~Rao, P.~Barnes, Y.~Tay, N.~Shazeer, V.~Prabhakaran, E.~Reif,
  N.~Du, B.~Hutchinson, R.~Pope, J.~Bradbury, J.~Austin, M.~Isard,
  G.~Gur{-}Ari, P.~Yin, T.~Duke, A.~Levskaya, S.~Ghemawat, S.~Dev,
  H.~Michalewski, X.~Garcia, V.~Misra, K.~Robinson, L.~Fedus, D.~Zhou,
  D.~Ippolito, D.~Luan, H.~Lim, B.~Zoph, A.~Spiridonov, R.~Sepassi, D.~Dohan,
  S.~Agrawal, M.~Omernick, A.~M. Dai, T.~S. Pillai, M.~Pellat, A.~Lewkowycz,
  E.~Moreira, R.~Child, O.~Polozov, K.~Lee, Z.~Zhou, X.~Wang, B.~Saeta,
  M.~Diaz, O.~Firat, M.~Catasta, J.~Wei, K.~Meier{-}Hellstern, D.~Eck, J.~Dean,
  S.~Petrov, and N.~Fiedel.
\newblock Palm: Scaling language modeling with pathways.
\newblock \emph{CoRR}, abs/2204.02311, 2022.

\bibitem[Chung et~al.(2022)Chung, Hou, Longpre, Zoph, Tay, Fedus, Li, Wang,
  Dehghani, Brahma, Webson, Gu, Dai, Suzgun, Chen, Chowdhery, Narang, Mishra,
  Yu, Zhao, Huang, Dai, Yu, Petrov, Chi, Dean, Devlin, Roberts, Zhou, Le, and
  Wei]{flant5}
H.~W. Chung, L.~Hou, S.~Longpre, B.~Zoph, Y.~Tay, W.~Fedus, E.~Li, X.~Wang,
  M.~Dehghani, S.~Brahma, A.~Webson, S.~S. Gu, Z.~Dai, M.~Suzgun, X.~Chen,
  A.~Chowdhery, S.~Narang, G.~Mishra, A.~Yu, V.~Y. Zhao, Y.~Huang, A.~M. Dai,
  H.~Yu, S.~Petrov, E.~H. Chi, J.~Dean, J.~Devlin, A.~Roberts, D.~Zhou, Q.~V.
  Le, and J.~Wei.
\newblock Scaling instruction-finetuned language models.
\newblock \emph{CoRR}, abs/2210.11416, 2022.

\bibitem[Devlin et~al.(2019)Devlin, Chang, Lee, and Toutanova]{bert}
J.~Devlin, M.~Chang, K.~Lee, and K.~Toutanova.
\newblock {BERT:} pre-training of deep bidirectional transformers for language
  understanding.
\newblock In \emph{{NAACL-HLT} {(1)}}, pages 4171--4186. Association for
  Computational Linguistics, 2019.

\bibitem[Dosovitskiy et~al.(2021)Dosovitskiy, Beyer, Kolesnikov, Weissenborn,
  Zhai, Unterthiner, Dehghani, Minderer, Heigold, Gelly, Uszkoreit, and
  Houlsby]{vit}
A.~Dosovitskiy, L.~Beyer, A.~Kolesnikov, D.~Weissenborn, X.~Zhai,
  T.~Unterthiner, M.~Dehghani, M.~Minderer, G.~Heigold, S.~Gelly, J.~Uszkoreit,
  and N.~Houlsby.
\newblock An image is worth 16x16 words: Transformers for image recognition at
  scale.
\newblock In \emph{{ICLR}}. OpenReview.net, 2021.

\bibitem[Driess et~al.(2023)Driess, Xia, Sajjadi, Lynch, Chowdhery, Ichter,
  Wahid, Tompson, Vuong, Yu, Huang, Chebotar, Sermanet, Duckworth, Levine,
  Vanhoucke, Hausman, Toussaint, Greff, Zeng, Mordatch, and Florence]{palm-e}
D.~Driess, F.~Xia, M.~S.~M. Sajjadi, C.~Lynch, A.~Chowdhery, B.~Ichter,
  A.~Wahid, J.~Tompson, Q.~Vuong, T.~Yu, W.~Huang, Y.~Chebotar, P.~Sermanet,
  D.~Duckworth, S.~Levine, V.~Vanhoucke, K.~Hausman, M.~Toussaint, K.~Greff,
  A.~Zeng, I.~Mordatch, and P.~Florence.
\newblock Palm-e: An embodied multimodal language model.
\newblock \emph{CoRR}, abs/2303.03378, 2023.

\bibitem[Hu et~al.(2022)Hu, Shen, Wallis, Allen{-}Zhu, Li, Wang, Wang, and
  Chen]{lora}
E.~J. Hu, Y.~Shen, P.~Wallis, Z.~Allen{-}Zhu, Y.~Li, S.~Wang, L.~Wang, and
  W.~Chen.
\newblock Lora: Low-rank adaptation of large language models.
\newblock In \emph{{ICLR}}. OpenReview.net, 2022.

\bibitem[Kudo and Richardson(2018)]{sentencepiece}
T.~Kudo and J.~Richardson.
\newblock Sentencepiece: {A} simple and language independent subword tokenizer
  and detokenizer for neural text processing.
\newblock In \emph{{EMNLP} (Demonstration)}, pages 66--71. Association for
  Computational Linguistics, 2018.

\bibitem[Li et~al.(2022)Li, Xu, Tian, Wang, Yan, Bi, Ye, Chen, Xu, Cao, Zhang,
  Huang, Huang, Zhou, and Si]{mplug}
C.~Li, H.~Xu, J.~Tian, W.~Wang, M.~Yan, B.~Bi, J.~Ye, H.~Chen, G.~Xu, Z.~Cao,
  J.~Zhang, S.~Huang, F.~Huang, J.~Zhou, and L.~Si.
\newblock mplug: Effective and efficient vision-language learning by
  cross-modal skip-connections.
\newblock In \emph{{EMNLP}}, pages 7241--7259. Association for Computational
  Linguistics, 2022.

\bibitem[Li et~al.(2023)Li, Li, Savarese, and Hoi]{blip2}
J.~Li, D.~Li, S.~Savarese, and S.~C.~H. Hoi.
\newblock {BLIP-2:} bootstrapping language-image pre-training with frozen image
  encoders and large language models.
\newblock \emph{CoRR}, abs/2301.12597, 2023.

\bibitem[Liu et~al.(2023)Liu, Li, Wu, and Lee]{llava}
H.~Liu, C.~Li, Q.~Wu, and Y.~J. Lee.
\newblock Visual instruction tuning.
\newblock \emph{CoRR}, abs/2304.08485, 2023.

\bibitem[OpenAI(2022)]{chatgpt}
OpenAI.
\newblock Introducing chatgpt.
\newblock \url{https://openai.com/blog/chatgpt}, 2022.

\bibitem[OpenAI(2023)]{gpt4}
OpenAI.
\newblock {GPT-4} technical report.
\newblock \emph{CoRR}, abs/2303.08774, 2023.

\bibitem[Ouyang et~al.(2022)Ouyang, Wu, Jiang, Almeida, Wainwright, Mishkin,
  Zhang, Agarwal, Slama, Ray, Schulman, Hilton, Kelton, Miller, Simens, Askell,
  Welinder, Christiano, Leike, and Lowe]{instructgpt}
L.~Ouyang, J.~Wu, X.~Jiang, D.~Almeida, C.~L. Wainwright, P.~Mishkin, C.~Zhang,
  S.~Agarwal, K.~Slama, A.~Ray, J.~Schulman, J.~Hilton, F.~Kelton, L.~Miller,
  M.~Simens, A.~Askell, P.~Welinder, P.~F. Christiano, J.~Leike, and R.~Lowe.
\newblock Training language models to follow instructions with human feedback.
\newblock \emph{CoRR}, abs/2203.02155, 2022.

\bibitem[Radford and Narasimhan(2018)]{gpt1}
A.~Radford and K.~Narasimhan.
\newblock Improving language understanding by generative pre-training.
\newblock 2018.

\bibitem[Raffel et~al.(2020)Raffel, Shazeer, Roberts, Lee, Narang, Matena,
  Zhou, Li, and Liu]{t5}
C.~Raffel, N.~Shazeer, A.~Roberts, K.~Lee, S.~Narang, M.~Matena, Y.~Zhou,
  W.~Li, and P.~J. Liu.
\newblock Exploring the limits of transfer learning with a unified text-to-text
  transformer.
\newblock \emph{J. Mach. Learn. Res.}, 21:\penalty0 140:1--140:67, 2020.

\bibitem[Scao et~al.(2022)Scao, Fan, Akiki, Pavlick, Ilic, Hesslow,
  Castagn{\'{e}}, Luccioni, Yvon, Gall{\'{e}}, Tow, Rush, Biderman, Webson,
  Ammanamanchi, Wang, Sagot, Muennighoff, del Moral, Ruwase, Bawden, Bekman,
  McMillan{-}Major, Beltagy, Nguyen, Saulnier, Tan, Suarez, Sanh,
  Lauren{\c{c}}on, Jernite, Launay, Mitchell, Raffel, Gokaslan, Simhi, Soroa,
  Aji, Alfassy, Rogers, Nitzav, Xu, Mou, Emezue, Klamm, Leong, van Strien,
  Adelani, and et~al.]{bloom}
T.~L. Scao, A.~Fan, C.~Akiki, E.~Pavlick, S.~Ilic, D.~Hesslow,
  R.~Castagn{\'{e}}, A.~S. Luccioni, F.~Yvon, M.~Gall{\'{e}}, J.~Tow, A.~M.
  Rush, S.~Biderman, A.~Webson, P.~S. Ammanamanchi, T.~Wang, B.~Sagot,
  N.~Muennighoff, A.~V. del Moral, O.~Ruwase, R.~Bawden, S.~Bekman,
  A.~McMillan{-}Major, I.~Beltagy, H.~Nguyen, L.~Saulnier, S.~Tan, P.~O.
  Suarez, V.~Sanh, H.~Lauren{\c{c}}on, Y.~Jernite, J.~Launay, M.~Mitchell,
  C.~Raffel, A.~Gokaslan, A.~Simhi, A.~Soroa, A.~F. Aji, A.~Alfassy, A.~Rogers,
  A.~K. Nitzav, C.~Xu, C.~Mou, C.~Emezue, C.~Klamm, C.~Leong, D.~van Strien,
  D.~I. Adelani, and et~al.
\newblock {BLOOM:} {A} 176b-parameter open-access multilingual language model.
\newblock \emph{CoRR}, abs/2211.05100, 2022.

\bibitem[Schuhmann et~al.(2021)Schuhmann, Vencu, Beaumont, Kaczmarczyk, Mullis,
  Katta, Coombes, Jitsev, and Komatsuzaki]{laion400m}
C.~Schuhmann, R.~Vencu, R.~Beaumont, R.~Kaczmarczyk, C.~Mullis, A.~Katta,
  T.~Coombes, J.~Jitsev, and A.~Komatsuzaki.
\newblock {LAION-400M:} open dataset of clip-filtered 400 million image-text
  pairs.
\newblock \emph{CoRR}, abs/2111.02114, 2021.

\bibitem[Sharma et~al.(2018)Sharma, Ding, Goodman, and Soricut]{conceptualcap}
P.~Sharma, N.~Ding, S.~Goodman, and R.~Soricut.
\newblock Conceptual captions: {A} cleaned, hypernymed, image alt-text dataset
  for automatic image captioning.
\newblock In \emph{{ACL} {(1)}}, pages 2556--2565. Association for
  Computational Linguistics, 2018.

\bibitem[Shen et~al.(2023)Shen, Song, Tan, Li, Lu, and Zhuang]{hugginggpt}
Y.~Shen, K.~Song, X.~Tan, D.~Li, W.~Lu, and Y.~Zhuang.
\newblock Hugginggpt: Solving {AI} tasks with chatgpt and its friends in
  huggingface.
\newblock \emph{CoRR}, abs/2303.17580, 2023.

\bibitem[Taori et~al.(2023)Taori, Gulrajani, Zhang, Dubois, Li, Guestrin,
  Liang, and Hashimoto]{alpaca}
R.~Taori, I.~Gulrajani, T.~Zhang, Y.~Dubois, X.~Li, C.~Guestrin, P.~Liang, and
  T.~B. Hashimoto.
\newblock Stanford alpaca: An instruction-following llama model.
\newblock \url{https://github.com/tatsu-lab/stanford_alpaca}, 2023.

\bibitem[Touvron et~al.(2023)Touvron, Lavril, Izacard, Martinet, Lachaux,
  Lacroix, Rozi{\`{e}}re, Goyal, Hambro, Azhar, Rodriguez, Joulin, Grave, and
  Lample]{llama}
H.~Touvron, T.~Lavril, G.~Izacard, X.~Martinet, M.~Lachaux, T.~Lacroix,
  B.~Rozi{\`{e}}re, N.~Goyal, E.~Hambro, F.~Azhar, A.~Rodriguez, A.~Joulin,
  E.~Grave, and G.~Lample.
\newblock Llama: Open and efficient foundation language models.
\newblock \emph{CoRR}, abs/2302.13971, 2023.

\bibitem[Vicuna(2023)]{vicuna}
Vicuna.
\newblock Vicuna: An open chatbot impressing gpt-4.
\newblock \url{https://github.com/lm-sys/FastChat}, 2023.

\bibitem[Wang et~al.(2022)Wang, Kordi, Mishra, Liu, Smith, Khashabi, and
  Hajishirzi]{self-instruct}
Y.~Wang, Y.~Kordi, S.~Mishra, A.~Liu, N.~A. Smith, D.~Khashabi, and
  H.~Hajishirzi.
\newblock Self-instruct: Aligning language model with self generated
  instructions.
\newblock \emph{CoRR}, abs/2212.10560, 2022.
\newblock \doi{10.48550/arXiv.2212.10560}.
\newblock URL \url{https://doi.org/10.48550/arXiv.2212.10560}.

\bibitem[Wu et~al.(2023)Wu, Yin, Qi, Wang, Tang, and Duan]{visualchatgpt}
C.~Wu, S.~Yin, W.~Qi, X.~Wang, Z.~Tang, and N.~Duan.
\newblock Visual chatgpt: Talking, drawing and editing with visual foundation
  models.
\newblock \emph{CoRR}, abs/2303.04671, 2023.

\bibitem[Xu et~al.(2023{\natexlab{a}})Xu, Guo, Duan, and McAuley]{baize}
C.~Xu, D.~Guo, N.~Duan, and J.~J. McAuley.
\newblock Baize: An open-source chat model with parameter-efficient tuning on
  self-chat data.
\newblock \emph{CoRR}, abs/2304.01196, 2023{\natexlab{a}}.

\bibitem[Xu et~al.(2021)Xu, Yan, Li, Bi, Huang, Xiao, and Huang]{e2evlp}
H.~Xu, M.~Yan, C.~Li, B.~Bi, S.~Huang, W.~Xiao, and F.~Huang.
\newblock {E2E-VLP:} end-to-end vision-language pre-training enhanced by visual
  learning.
\newblock In \emph{{ACL/IJCNLP} {(1)}}, pages 503--513. Association for
  Computational Linguistics, 2021.

\bibitem[Xu et~al.(2023{\natexlab{b}})Xu, Ye, Yan, Shi, Ye, Xu, Li, Bi, Qian,
  Wang, Xu, Zhang, Huang, Huang, and Zhou]{mplug2}
H.~Xu, Q.~Ye, M.~Yan, Y.~Shi, J.~Ye, Y.~Xu, C.~Li, B.~Bi, Q.~Qian, W.~Wang,
  G.~Xu, J.~Zhang, S.~Huang, F.~Huang, and J.~Zhou.
\newblock mplug-2: {A} modularized multi-modal foundation model across text,
  image and video.
\newblock \emph{CoRR}, abs/2302.00402, 2023{\natexlab{b}}.

\bibitem[Yang et~al.(2023)Yang, Li, Wang, Lin, Azarnasab, Ahmed, Liu, Liu,
  Zeng, and Wang]{mmreact}
Z.~Yang, L.~Li, J.~Wang, K.~Lin, E.~Azarnasab, F.~Ahmed, Z.~Liu, C.~Liu,
  M.~Zeng, and L.~Wang.
\newblock {MM-REACT:} prompting chatgpt for multimodal reasoning and action.
\newblock \emph{CoRR}, abs/2303.11381, 2023.

\bibitem[Ye et~al.(2022)Ye, Xu, Yan, Xu, Qian, Zhang, and Huang]{hitea}
Q.~Ye, G.~Xu, M.~Yan, H.~Xu, Q.~Qian, J.~Zhang, and F.~Huang.
\newblock Hitea: Hierarchical temporal-aware video-language pre-training.
\newblock \emph{CoRR}, abs/2212.14546, 2022.
\newblock \doi{10.48550/arXiv.2212.14546}.
\newblock URL \url{https://doi.org/10.48550/arXiv.2212.14546}.

\bibitem[Zhang et~al.(2022)Zhang, Roller, Goyal, Artetxe, Chen, Chen, Dewan,
  Diab, Li, Lin, Mihaylov, Ott, Shleifer, Shuster, Simig, Koura, Sridhar, Wang,
  and Zettlemoyer]{opt}
S.~Zhang, S.~Roller, N.~Goyal, M.~Artetxe, M.~Chen, S.~Chen, C.~Dewan, M.~T.
  Diab, X.~Li, X.~V. Lin, T.~Mihaylov, M.~Ott, S.~Shleifer, K.~Shuster,
  D.~Simig, P.~S. Koura, A.~Sridhar, T.~Wang, and L.~Zettlemoyer.
\newblock {OPT:} open pre-trained transformer language models.
\newblock \emph{CoRR}, abs/2205.01068, 2022.

\bibitem[Zhu et~al.(2023{\natexlab{a}})Zhu, Chen, Shen, Li, and
  Elhoseiny]{minigpt4}
D.~Zhu, J.~Chen, X.~Shen, X.~Li, and M.~Elhoseiny.
\newblock Minigpt-4: Enhancing vision-language understanding with advanced
  large language models, 2023{\natexlab{a}}.

\bibitem[Zhu et~al.(2023{\natexlab{b}})Zhu, Hessel, Awadalla, Gadre, Dodge,
  Fang, Yu, Schmidt, Wang, and Choi]{openflamingo}
W.~Zhu, J.~Hessel, A.~Awadalla, S.~Y. Gadre, J.~Dodge, A.~Fang, Y.~Yu,
  L.~Schmidt, W.~Y. Wang, and Y.~Choi.
\newblock Multimodal {C4:} an open, billion-scale corpus of images interleaved
  with text.
\newblock \emph{CoRR}, abs/2304.06939, 2023{\natexlab{b}}.

\end{thebibliography}
%%%%%%%%%%%%%%%%%%%%%%%%%%%%%%%%%%%%%%%%%%%%%%%%%%%%%%%%%%%%

\clearpage
% Appendix
\newpage
\appendix
\section{Training Hyperparameters}
We report the detailed model training hyperparameters for visual knowledge learning in Table~\ref{tbl:hyperparam:pt} and vision-language joint instruction tuning in Table~\ref{tbl:hyperparam:ft}.

\begin{table}[ht]
\centering
\begin{tabular}{lc}
\toprule
\textbf{Hyperparameters} & \\ \midrule
Training steps             &       50,000 \\
Warmup steps                      &       375 \\
Max length        &       512 \\
Batch size of image-caption pairs  &       4,096 \\
Optimizer & AdamW \\
Learning rate & 2e-4 \\
Learning rate decay & Cosine \\
Adam $\epsilon$ & 1e-6 \\
Adam $\beta$ & (0.9, 0.98) \\
Weight decay & 0.01 \\
\bottomrule
\end{tabular}
\vspace{1ex}
\caption{Training hyperparameters for multi-modal pre-training stage.}
\label{tbl:hyperparam:pt}
\end{table}

\begin{table}[ht]
\centering
\begin{tabular}{lc}
\toprule
\textbf{Hyperparameters} & \\ \midrule
Training steps                   &       2,000 \\
Warmup steps                      &       50 \\
Max length         &       1,024 \\
Batch size of text instruction data  &       128 \\
Batch size of multi-modal instruction data  &   128 \\
Optimizer & AdamW \\
Learning rate & 2e-5 \\
Learning rate decay & Cosine \\
AdamW $\epsilon$ & 1e-6 \\
AdamW $\beta$ & (0.9, 0.999) \\
Weight decay & 0.0001 \\
\bottomrule
\end{tabular}
\vspace{1ex}
\caption{Training hyperparameters for vision-language joint instruction tuning stage.}
\label{tbl:hyperparam:ft}
\end{table}

\section{Comparison with MM-REACT}
\begin{figure}[!ht]
     \centering
     \includegraphics[width=0.7 \textwidth]{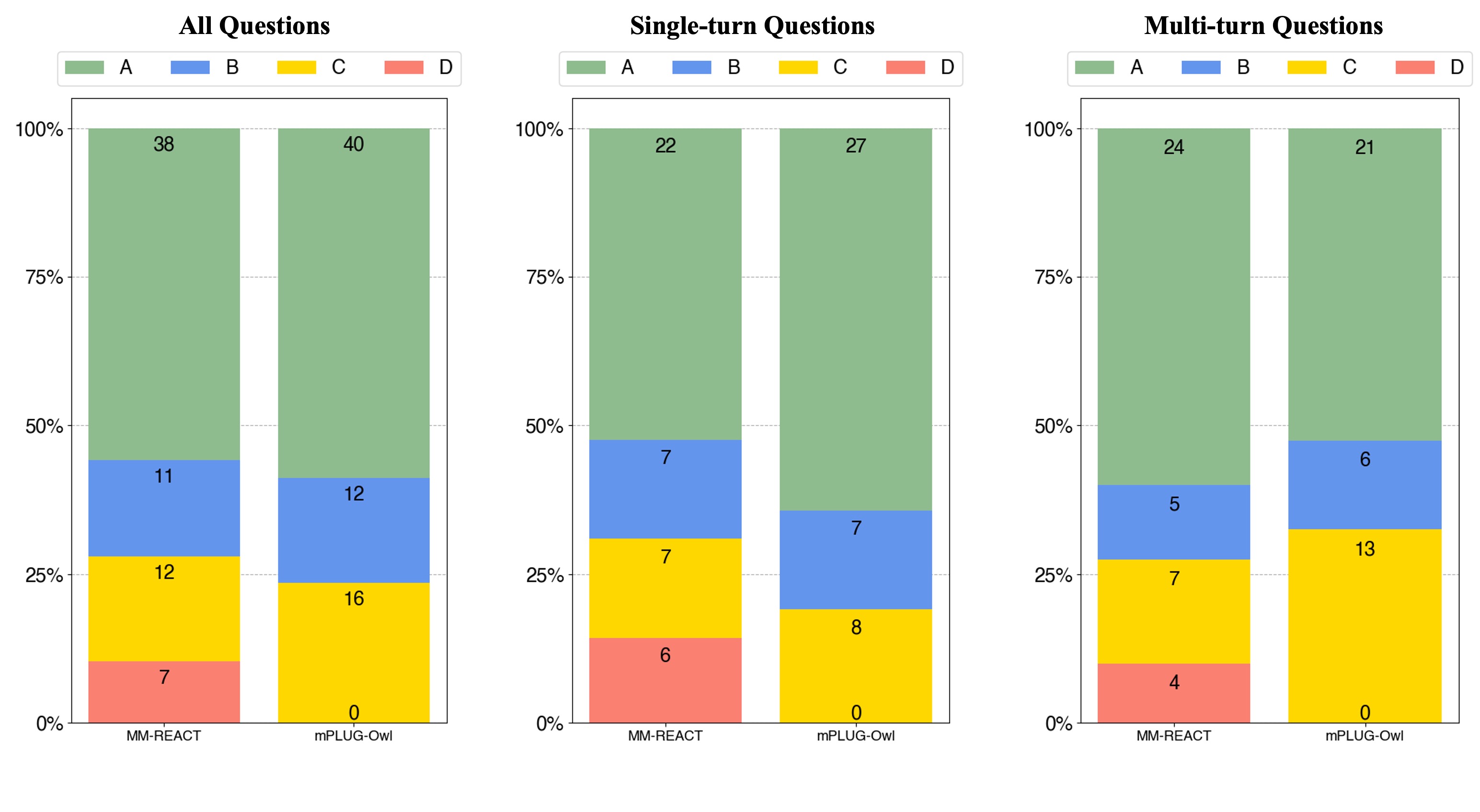}
     \caption{The comparison results which exclude the cases that were generated unsuccessfully by MM-REACT.}
     \label{fig:mm-react}
     \vspace{-2mm}
\end{figure}

\begin{figure}[!ht]
    \centering
    \includegraphics[width=1 \textwidth]{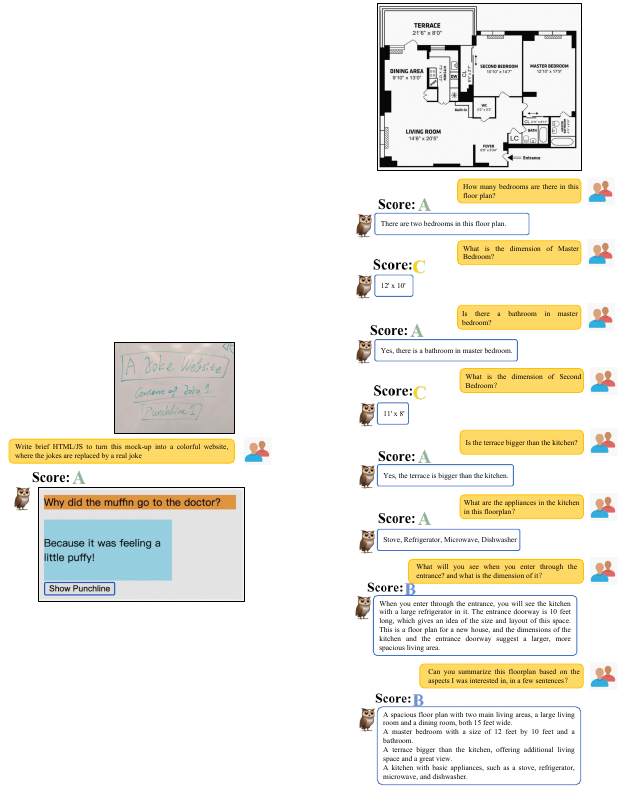}
    \caption{OCR of simple scenes (mostly scenes with few numbers and no calculation a).}
    \label{fig:OCR_1_scoreB}
    \vspace{-2mm}
\end{figure}

\begin{figure}[!ht]
    \centering
    \includegraphics[width=1 \textwidth]{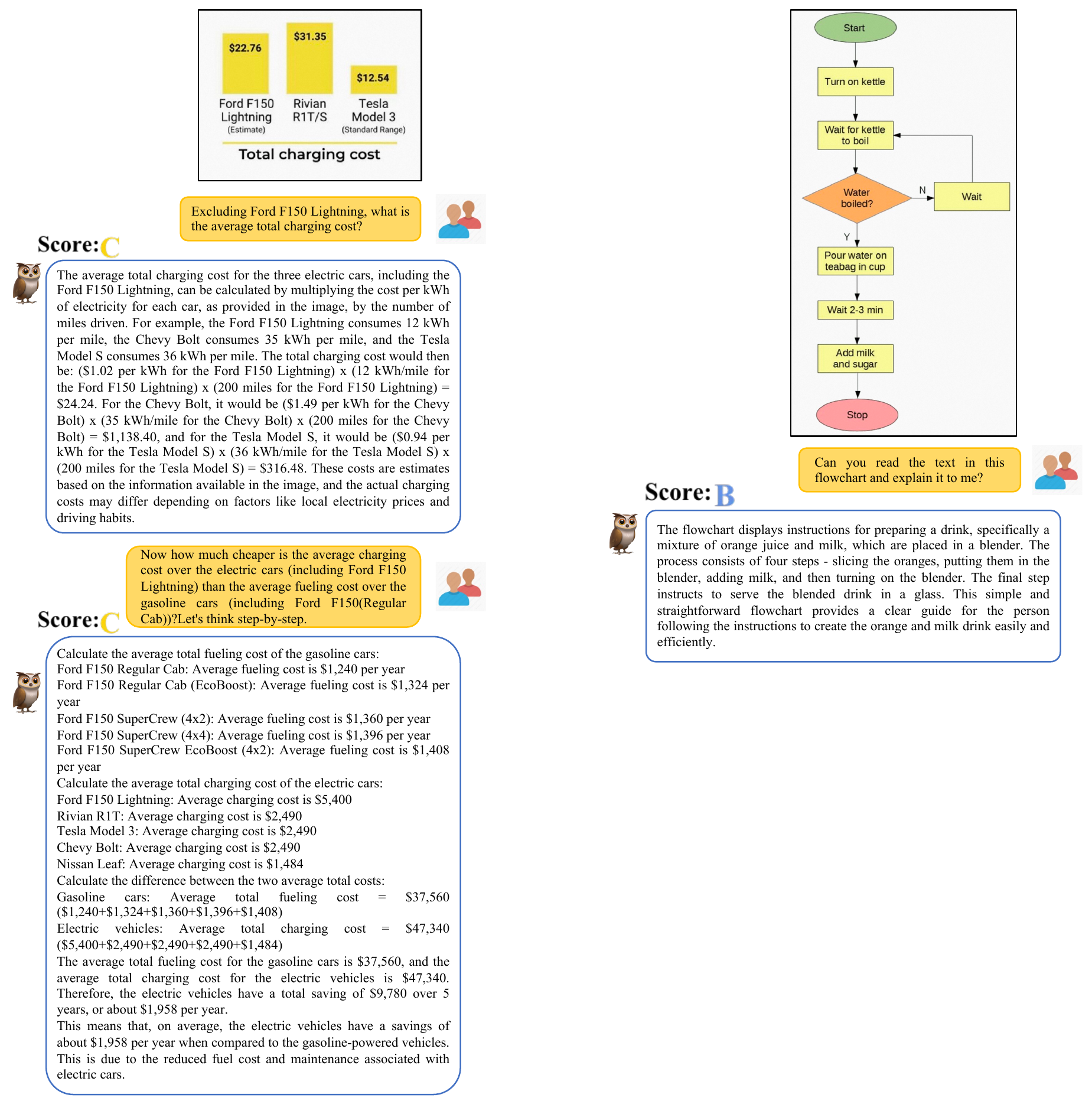}
    \caption{OCR of complex scenes (a).}
    \label{fig:OCR_1_scoreC_a}
    \vspace{-2mm}
\end{figure}

\begin{figure}[!ht]
    \centering
    \includegraphics[width=\textwidth]{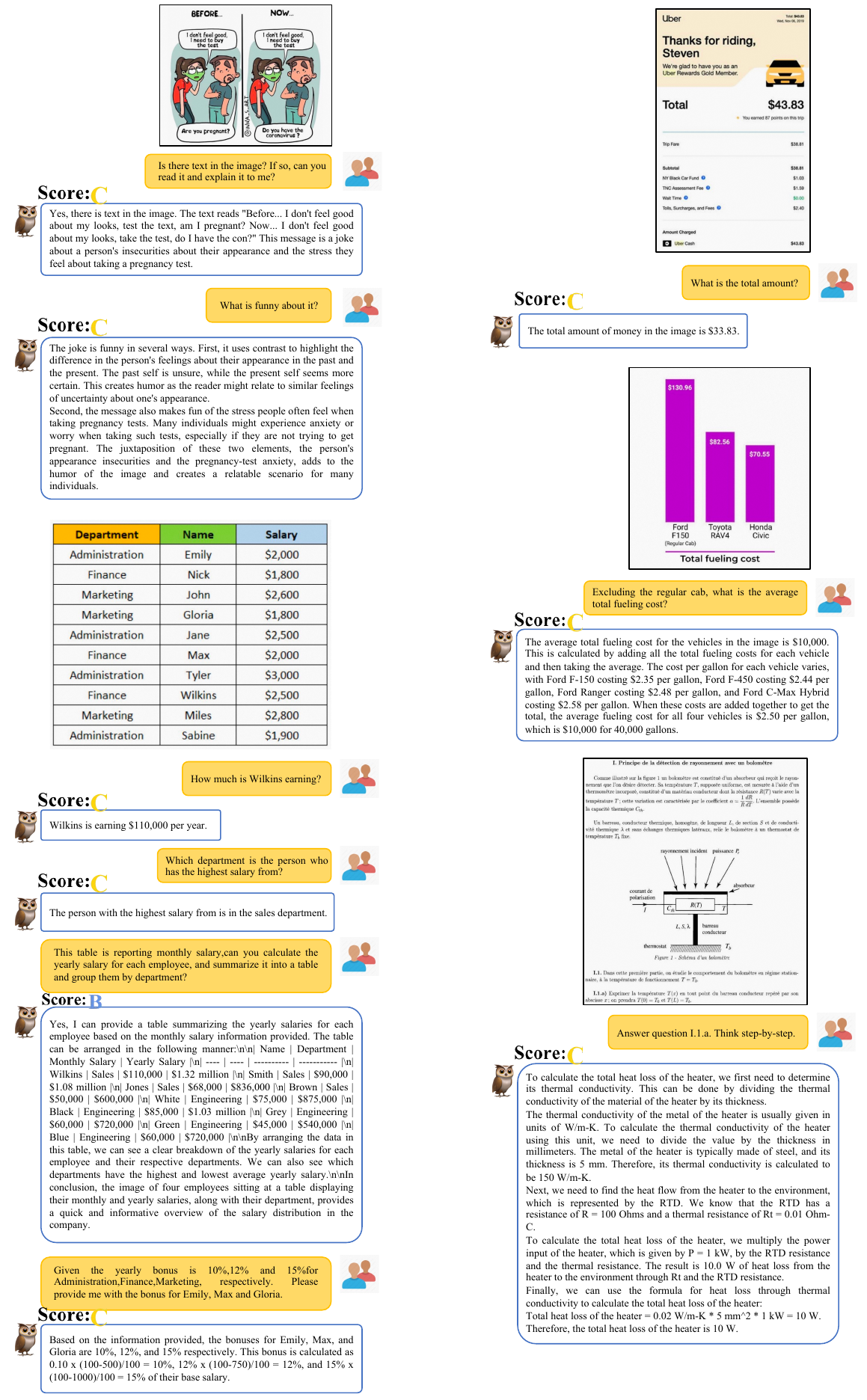}
    \caption{OCR of complex scenes (b).}
    \label{fig:OCR_1_scoreC_b}
    \vspace{-2mm}
\end{figure}

\end{document}